%% file: main.tex
\documentclass[sigconf]{acmart}

\usepackage{import}
\usepackage{example}
\usepackage{tikz}
\usepackage{arydshln}
\usepackage{pgfplots}
\pgfplotsset{compat=1.18} 
\usepackage{adjustbox}
\usepackage{listings}
\usetikzlibrary{shadows.blur}
\usepackage{tikzpagenodes}
\usepackage[utf8]{inputenc}
\usepackage{listings}
\usepackage{amsmath}
\usepackage{algorithm}
\usepackage[noend]{algpseudocode}
\usepackage{hyperref}
\usepackage[vietnamese=nohyphenation]{hyphsubst}
\usepackage[vietnamese,english]{babel}
\usepackage[figurename=Fig.]{caption}
\usepackage{makecell}
\usepackage{subfigure}
\usepackage{graphicx}
\usepackage{natbib} %

\AtBeginDocument{%
  \providecommand\BibTeX{{%
    \normalfont B\kern-0.5em{\scshape i\kern-0.25em b}\kern-0.8em\TeX}}}

\begin{document}

\title{Cross-Data Knowledge Graph Construction for LLM-enabled Educational Question-Answering System: A~Case~Study~at~HCMUT}

\author{Tuan Bui}
\authornotemark[1]
\email{tuanbc88@hcmut.edu.vn}
\orcid{0000-0002-8587-182X}
\affiliation{%
  \institution{Ho Chi Minh City University of Technology (HCMUT)}
  \city{HoChiMinhCity}
  \country{VietNam}
}

\author{Oanh Tran}
\email{oanh.tranotsc1123@hcmut.edu.vn}
\affiliation{%
  \institution{Ho Chi Minh City University of Technology (HCMUT)}
  \city{HoChiMinhCity}
  \country{VietNam}
}

\author{Phuong Nguyen}
\email{phuong.nguyenvoid@hcmut.edu.vn}
\affiliation{%
  \institution{Ho Chi Minh City University of Technology (HCMUT)}
  \city{HoChiMinhCity}
  \country{VietNam}
}

\author{Bao Ho}
\email{bao.ho64qubit@hcmut.edu.vn}
\affiliation{%
  \institution{Ho Chi Minh City University of Technology (HCMUT)}
  \city{HoChiMinhCity}
  \country{VietNam}
}

\author{Long Nguyen}
\email{long.nguyencse2023@hcmut.edu.vn}
\affiliation{%
  \institution{Ho Chi Minh City University of Technology (HCMUT)}
  \city{HoChiMinhCity}
  \country{VietNam}
}

\author{Thang Bui}
\authornotemark[2]
\email{bhthang@hcmut.edu.vn}
\affiliation{%
  \institution{Ho Chi Minh City University of Technology (HCMUT)}
  \city{HoChiMinhCity}
  \country{VietNam}
}

\author{Tho Quan}
\authornotemark[2]
\email{qttho@hcmut.edu.vn}
\orcid{0000−0003−0467−6254}

\affiliation{%
  \institution{Ho Chi Minh City University of Technology (HCMUT)}
  \city{HoChiMinhCity}
  \country{VietNam}
}

\renewcommand{\shortauthors}{Tuan Bui, et al.}

\begin{abstract}
In today's rapidly evolving landscape of Artificial Intelligence, large language models (LLMs) have emerged as a vibrant research topic. LLMs find applications in various fields and contribute significantly. Despite their powerful language capabilities, similar to pre-trained language models (PLMs), LLMs still face challenges in remembering events, incorporating new information, and addressing domain-specific issues or hallucinations. To overcome these limitations, researchers have proposed Retrieval-Augmented Generation (RAG) techniques, some others have proposed the integration of LLMs with Knowledge Graphs (KGs) to provide factual context, thereby improving performance and delivering more accurate feedback to user queries.

Education plays a crucial role in human development and progress. With the technology transformation, traditional education is being replaced by digital or blended education. Therefore, educational data in the digital environment is increasing day by day. Data in higher education institutions are diverse, comprising various sources such as unstructured/structured text, relational databases, web/app-based API access, etc. Constructing a Knowledge Graph from these cross-data sources is not a simple task. This article proposes a method for automatically constructing a Knowledge Graph from multiple data sources and discusses some initial applications (experimental trials) of KG in conjunction with LLMs for question-answering tasks.
  
\end{abstract}





\maketitle

\section{Introduction}

A \textit{Large Language Model} (LLM) is a \textit{language model} trained on a considerably large size corpus. Nowadays, LLMs attract much attention from various communities due to their capacity to accomplish versatile language generation and comprehension. 
Notable LLM models reported to the public include OpenAI's GPT \cite{radford2018improving}, Google's BART \cite{lewis2019bart}, LLaMa \cite{touvron2023llama2} . To achieve a broad spectrum of general knowledge \cite{petroni2019language} and refine their language ability \cite{wei2022emergent}, state-of-the-art LLMs nowadays are trained on a massive corpus of textual data. 
However, an LLM's completion might contain "hallucinations"
\cite{Ji_2023} 
because of limited access to information that is as up-to-date, proprietary, or domain-specific as humans. To address this issue and other limitations, some hybrid models have shown prominent results by combining parametric memory with non-parametric memory \cite{karpukhin2020dense}. 
Since their knowledge bases can be directly revised and expanded, and retrieved knowledge can be inspected and interpreted. This technique, which received a lot of traction after the publication of the famous Facebook research paper \cite{lewis2021retrievalaugmented}, is well-known and coined as \textit{Retrieval-Augmented Generation (RAG)}. 


However, in most real-world domains such as education, the data/information often originate from various departments and faculties within an institution, resulting in diverse data sources with different natures such as structured text, unstructured text, databases, images, or access through API mechanisms from existing web/app-based systems. For instance, at \textit{Ho Chi Minh City University of Technology (HCMUT)}\footnote{https://hcmut.edu.vn}, unstructured data originate from legal documents, frequently asked questions (FAQs) from student support systems BKSI\footnote{https://mybk.hcmut.edu.vn/bksi/public/vi/} (Help Desk System), news from the website, data from databases, and information retrieved from API-enabled systems like teaching management systems LMS\footnote{https://lms.hcmut.edu.vn/}. Due to the cross-data nature, \textit{Knowledge Graphs (KGs)}  \cite{Hogan_2021} serve as a suitable non-parametric memory formalism for knowledge representation in the educational environment, allowing the effective deployment of a RAG system in this context. However, constructing a Knowledge Graph from multiple data sources, not specifically designed to be interoperable, is not a trivial task, in particular when one needs to deal with \textit{open intents} [ref] commonly arising during casual conversations between students and the university staffs. Hence, to the best of our knowledge, there has not been a best-practice application of RAG from KG for LLMs in practical scenarios.

In this paper, we aim to pioneer a KG-based RAG approach for an Educational Question-Answering system. We propose a framework for the cross-data knowledge graph construction in the educational domain, which is currently implemented at HCMUT and leverages the Vietnamese language. This framework is applied as a pilot in a Language Model (LLM)-based system. Our contributions are of three-fold: (i) We introduce the technique of intental entity discovery for unstructured text in Vietnamese FAQ conversations; (ii) we present the \textit{embedding-based cross-data} for relation discovery on education KG construction; and (iii) we conduct real-world experiments, specifically, LLM-enabled KG-based Question-Answering at HCMUT using RAG with the constructed educational KG.

\section{Related Work}
\subsection{Open Intent Discovery}
The task of \textit{Open Intent Discovery} \cite{Chen2022IntentDF} presents several challenges and difficulties in the field of \textit{Natural Language Understanding} and dialogue systems. One significant challenge is the inherent ambiguity and variability in user expressions. Unlike traditional intent recognition, where a predefined set of categories is used for classification, open intent discovery involves identifying user intents that may not have been encountered during the training phase. This introduces a level of uncertainty and unpredictability, as users can express their intents in diverse and unanticipated ways. Another challenge is the lack of sufficient labeled data for training models in an open-world setting. Creating labeled datasets for every potential intent becomes impractical, especially when dealing with a wide range of domains and applications. This scarcity of annotated examples for novel intents makes it challenging for models to generalize effectively and accurately identify open intents. In recent years, there has been an increased interest in determining user intent from both written and spoken language, with a focus on modeling and comprehending interactions. Recent studies, such as a method employing a bidirectional LSTM and CRF for detecting user intent efficiently \cite{vedula2019open}, an unsupervised two-stage approach aimed at discovering intents and generating meaningful intent labels automatically from unlabeled utterances \cite{Vedula2020}, and a method mining unlabeled utterance data to uncover common intents \cite{liu2021open}. These studies collectively highlight the significance of robust approaches for understanding user intent, paving the way for enhanced dialogue systems and virtual assistants. Open intents has been studied in fields such as Customer Care in businesses \cite{liu2021open,Vedula2020} and Healthcare facilities\cite{mullick2023intent}; however, in the realm of education, particularly in Student Care, it remains relatively unexplored. 

\subsection{Knowledge Graph in Education Domain} 
Knowledge Graphs (KGs) have evolved as an effective way to represent knowledge. KGs present a structured and integrated representation of concepts, relationships, and attributes within a domain. There have been many studies on Educational KGs such as knowledge graph for mathematics \cite{Chen_20218}, ontology modeling university teaching programs and student profiles (EducOnto) \cite{Hubert_2022}, knowledge schema for university teaching \cite{Rizun_2019}, and knowledge graphs to identify labor market needs (education and employment) \cite{Fettach_2022}. Notably, research on KGs for the Vietnamese educational area is sparse. Huong and Phuc \cite{Huong2020} introduced a framework for extracting triples (subject-predicate-object relationships) from Vietnamese. However, their technique has limitations in handling complex sentences and is applied to the travel area.

\subsection{KG-augmented LLMs} 

One major hurdle for LLMs is their occasional struggle with factual accuracy and hallucinations. This is where KGs come in. KGs offer a structured representation of knowledge, encoding entities and their relationships in a machine-readable format. \textit{KG-augmented LLMs} aim to bridge this gap by leveraging the strengths of both approaches. If the generated information from LLMs aligns with the "ground truth" represented by the KGs, it can be considered factually accurate and non-hallucinatory. This enhancing method involves refining the inference process of LLMs \cite{baek-etal-2023-knowledge}, 
optimizing learning mechanisms 
\cite{kim-etal-2023-cot}
, and establishing a result validation mechanism
\cite{kang2023knowledge}.
Significant progress has been made through these studies, highlighting the importance of continuous innovation and supporting the advancement of more advanced KG-augmented LLMs.

\section{Towards Cross-Data Knowledge Graph Construction for LLM-enabled Educational
Question-Answering System}


In this section, we delve into the methodological approach employed in our research to extract valuable insights from a multi-source educational data environment for KG construction, served as the foundation of an LLM-enabled educational QA system. We begin by discussing the nature of the data sources in the university environment of HCMUT and how they contribute to a comprehensive understanding of the educational ecosystem. We then introduce the \textit{E-OED Framework (Educational Open Entity Discovery)} , a robust methodology designed to explore intents using unsupervised learning methods. This framework is particularly adept at handling challenges such as overlapping entities, open intents, and data fusion from multiple sources. Lastly, we present our method for Relation Discovery, allowing us to explore relationships among different types of entities and construct a KG representing the education domain.

\subsection{Multi-source educational data at HCMUT}
In the dynamic landscape of a university environment like HCMUT, data streams from a myriad of sources, forming a complex multi-source ecosystem. 
This diverse and interconnected web of entities provides a comprehensive picture of the university ecosystem, enabling a deeper understanding of user needs, educational processes, and the overall university environment. 
includes the following data sources.

Figure \ref{fig:figure_01_multisource_data} presents a cross-data environment at HCMUT, which includes the following data sources.

\begin{figure}[h]
\centering
\subimport{latex_figures/}{figure_01_multisource_data.tex}
\caption{Multi-source educational data at HCMUT}
\label{fig:figure_01_multisource_data}
\end{figure}


\begin{itemize}
  \item \textit{FAQ and Help Desk System}: In HCMUT, there is an online interactive system allowing students to raise their concerns arising during their study and feedback received from academic staff. Knowledge extractable from this source includes open intents, events, student issues, university services, the software system of the university.
  \item Academic Portals/Websites: These are all the websites that HCMUT updates frequently for academic news, university policies/regulations, and course information. Knowledge extractable from this source includes guidance documents, detailed curricula, and program descriptions offered by HCMUT, along with other entities such as organizations (university, faculty, department), locations (address number, city, district), and person names (students, lecturers, staffs).
  \item \textit{Learning Management System (LMS) and other supporting systems}: These systems help students keep track of their coursework throughout the school year. Knowledge extractable from this source includes courses, keywords, discipline terminology, discipline theory, books, documents, journals, and course participants (students, lecturers, teaching assistants).
\end{itemize}


\subsection{The E-OED Framework}

\begin{figure}[h]
\centering
\subimport{latex_figures/}{figure_02_education_knowledge_graph.tex}
\vspace{-0.9cm}\caption{Entity type in education domain}
\label{fig:figure_02_education_knowledge_graph}
\end{figure}

Figure \ref{fig:figure_02_education_knowledge_graph} depicts the intricate task of extracting relationships within the educational domain and exploring intents. For instance, a certain \textit{Student} may have his \textit{Student status}, which is managed by \textit{Academic affairs office}. This student also belongs to a certain \textit{Faculty} and is enrolling in various instances of \textit{Class}. The student can also leverage some academic service when necessary such as \textit{Course drop} or \textit{Transcript inquiry}.
The graph showcases the complexity of mapping intents to academic entities, illustrating the nuanced connections that exist within this domain. This complexity underscores the necessity for robust methodologies to accurately decipher and analyze such relationships. The most challenging task towards such a KG construction is perhaps \textit{educational open entity discovery}.
These challenges include handling overlapping entities, exploring open intents (which lack extensive research compared to classified intents), and fusing data from multiple sources. Moreover, the Vietnamese language presents additional hurdles due to its status as a low-resource language. Furthermore, within educational environments, the intent of conversations are dynamic rather than fixed, and opened rather than limited. Consequently, the identification of open entities is crucial in the educational sphere.
The \textit{E-OED Framework (Educational Open Entity Discovery)}, represented by Figure \ref{fig:figure_03_intent_discovery_pipeline}, aims to discover intents from FAQ data. This framework includes three processing modules known as \textit{Data Preprocessing}, \textit{Semantic Clustering}, and \textit{Automatic Cluster Labeling}, each of which plays a crucial role in the intent discovery process. 

\begin{figure}[h]
\centering
\subimport{latex_figures/}{figure_03_intent_discovery_pipeline.tex}
\vspace{-0.2cm}\caption{Educational Open Entity Discovery Framework}
\label{fig:figure_03_intent_discovery_pipeline}
\end{figure}


\textbf{Data Preprocessing} 
In this step, we split paragraphs to a list of single sentences,
filtering out the noise, etc. 
This initial filtering helps us focus on the meat of the text.

\textbf{Semantic Clustering}
To obtain sentence embedding, we employ SimCSE \cite{gao2022simcse} as our chosen embedding model. The model we use adheres to the SimCSE framework and has been developed utilizing the renowned pre-trained PhoBERT \cite{nguyen-tuan-nguyen-2020-phobert} base. Before applying the embeddings to the clustering algorithm, it is essential to conduct dimension reduction methods. This precautionary step is taken to address the "curse of dimensionality," which can adversely impact the accuracy of clustering algorithms by influencing distance metrics and introducing unintended noise. We opt for UMAP over PCA and LDA.  After implementing dimensionality reduction, we opt for HDBSCAN \cite{mcinnes2017hdbscan} as our clustering algorithm. 

\textbf{Automatic Cluster Labeling}
The final step in this process involves the automatic labeling of the clusters. To accomplish this task, our initial approach entails employing the word-segmentation tool for sentence segmentation within each cluster. Subsequently, the segmented sentences undergo analysis using the  \textit{PhoNLP} \cite{phonlp} to execute both POS tagging and dependency tagging. Following the extraction of tags by the PhoNLP, a rule-based approach is implemented. 
Further refinement is achieved by identifying the most frequently occurring tokens within each subset, which are then designated as our cluster labels. Figure \ref{fig:figure_07_list_intentions} presents a compilation of some of the discovered intents from the FAQ dataset as the results of \textit{Semantic Clustering} and \textit{Cluster Labelling} processes. Involving courses and class groups, there exists a multitude of inquiries ranging from course enrollment, class transfers, and capacity expansions, to timetable matters. Investigating the intents and associated entities can aid in providing good responses or initiating subsequent actions.

\begin{figure}[h]
\centering
\subimport{latex_figures/}{figure_07_list_intentions.tex}
\vspace{-0.5cm}\caption{List of some discovered intents}
\label{fig:figure_07_list_intentions}
\end{figure}


\subsection{Embedding-based method for Relation Discovery}
In the context of Relation Discovery between Intents and other entities, we adopt an Embedding-based approach. Due to resource and time constraints, we initially experimented with two major types of entities: intent and policy. To accommodate a large amount of diverse data with varying lengths, we employ two-stage retriever. We use Sentence-BERT to embed intent entities and policy entities.
Once the embeddings are generated, we map different types of entities by measuring similarity or proximity between entity embeddings in multidimensional space.

After getting mapping between entities using embedding, we use tf\_idf to rerank and give higher scores to policies that have similar keywords with the intents. 
\cite{guo2016entity}, we identify implicit connections between entities. This analysis enables us to uncover meaningful associations and dependencies that might not be readily apparent from raw data alone. Figure \ref{fig:figure_06_embedding_based_method} illustrates the approach of using Embedding-based methods for relation discovery. 
Based on the discovered relationships, we construct a Knowledge Graph (KG) representing the education domain. Each entity type is depicted as a node in the graph, and relationships between entities are depicted as edges. The KG offers a structured representation of the semantic landscape within the education domain, facilitating efficient traversal, query processing, and inference tasks.

\begin{figure}[h]
\centering
\subimport{latex_figures/}{figure_06_embedding_based_method.tex}
\vspace*{-0.55cm}\caption{Relation Discovery Module using Embedding-based method}
\label{fig:figure_06_embedding_based_method}
\end{figure}


\section{Experiments}
\subsection{Datasets}

The experimental datasets consist of three different datasets \textit{Banking77\_eng}, \textit{Banking77\_vni}, and \textit{FAQ\_HCMUT\_vni} presented in Table \ref{tab:intent_discovery_result}. Banking77\_en \cite{casanueva2020efficient} is an English dataset that contains 77 customer intents from over 10,000 questions in the banking domain. Banking77\_vni is a Vietnamese dataset that is an auto-translation (using Google Translate) of the Banking77\_eng dataset. FAQ\_HCMUT\_vni is a Vietnamese dataset that contains over 200,000 frequently asked questions (FAQs) collected from the HelpDesk system of the HCMUT.
The Banking77\_eng and Banking77\_vni datasets are used to evaluate the performance of the our framework. While FAQ\_HCMUT\_vni dataset is used to demonstrate the result of the OED framework. To investigate the impact of different embeddings on clustering performance, our framework utilizes Vietnamese SimCSE for Vietnamese datasets, while the original BERTopic employs a sentence-transformers model (all-miniLM-L6 variant). 


\subsection{Discovered Educational Open Intents}

Table \ref{tab:intent_discovery_result} illustrates the outcomes of our experimentation in discovering intents across the specified datasets. To validate the OED Framework, we executed the BERTopic framework on two datasets, Banking77\_eng and Banking77\_vni. The clustering results in Cases 1 and 2 demonstrate the superior performance of the BERTopic framework with the English dataset, where 73 intents were identified, closely aligning with the 77 predetermined categories. However, in the Banking77\_vni dataset, the count of identified intents notably decreased to 65. In Case 3, the OED Framework exhibited better performance with the Banking77\_vni dataset compared to the BERTopic framework for the Vietnamese language, yielding 76 extracted intents. Subsequently, we applied the OED Framework to the HCMUT\_FAQ\_vni dataset (Case 4), resulting in a total of 284 clusters with significant noised clusters, and duplicated clusters. Despite this, approximately 372 intents were derived from these clusters, although a substantial number of intents remain undiscovered. As previously mentioned, Figure  \ref{fig:figure_07_list_intentions} 
 illustrates some remarkable open intents discovered by our approach. 

\begin{table}[H]
  \caption{Open intent discovery result}
  \label{tab:intent_discovery_result}
\begin{adjustbox}{width=0.46\textwidth}
  \begin{tabular}{cllrr}
    \toprule
    \textbf{Case} & \textbf{Framework} & \textbf{Dataset} & \textbf{Cluster's No} & \textbf{Extracted intent} \\
    \midrule
    \texttt{1} & BERTopic & Banking77\_eng & 157 & 73 \\
    \texttt{2} & BERTopic &  Banking77\_vni & 147 & 65 \\
    \texttt{3} & OED Framework &  Banking77\_vni & 257 & \textbf{76}  \\
    \texttt{4} & OED Framework & HCMUT\_FAQ\_vni& 284 & \textbf{372}  \\
    \bottomrule
  \end{tabular}
  \end{adjustbox}
\end{table}



\subsection{Embedding-based method for Relation Discovery Result}

The embedding-based approach for relation discovery has yielded promising results in exploring relationships between entities, particularly in the education domain. Table \ref{tab:relation_discovery_result} show the result of Embedding-based approach, in this experiment, there are 243 intent entities and 237 policy entities. Following our approach, and after several trials, we've settled on a threshold of 0.32. A total of 613 relationships between entity pairs (intent and policy) have been identified. Out of these, 53 intent entities do not have any associations with policy entities. For instance, the intent "download form" is broad and lacks specific entity references. Additionally, among these 53 intent entities without relationships, 22 intents are overlooked, indicating they indeed have connections with policies but were not detected. For example, "Cancel course" and "Withdraw course" are linked to the Withdrawal Policy.

\begin{table}[H]
  \caption{Relation discovery result}
  \label{tab:relation_discovery_result}
  \begin{adjustbox}{width=0.25\textwidth}
  \begin{tabular}{lccc}
    \toprule
    \textbf{Entity pair}  & \textbf{[Intent, Policy]}    \\
    \midrule
    No of entities & [243 , 237]  \\
    Discovered relationships & \textbf{613}  \\
    Non-associative intents & 53   \\
    Overlooked intents & \textbf{22} \\ 
    \bottomrule
  \end{tabular}
  \end{adjustbox}
\end{table}

\subsection{KG-augmented LLMs Approach in Educational Context}

\begin{figure}[h!]
\centering
\subfigure[KG-augmented LLMs Approach]
{\subimport{latex_figures/}{figure_04_kgrag_pipeline.tex}}
\subfigure[Demonstration of course withdrawal case]
{\includegraphics[width=0.5\textwidth]{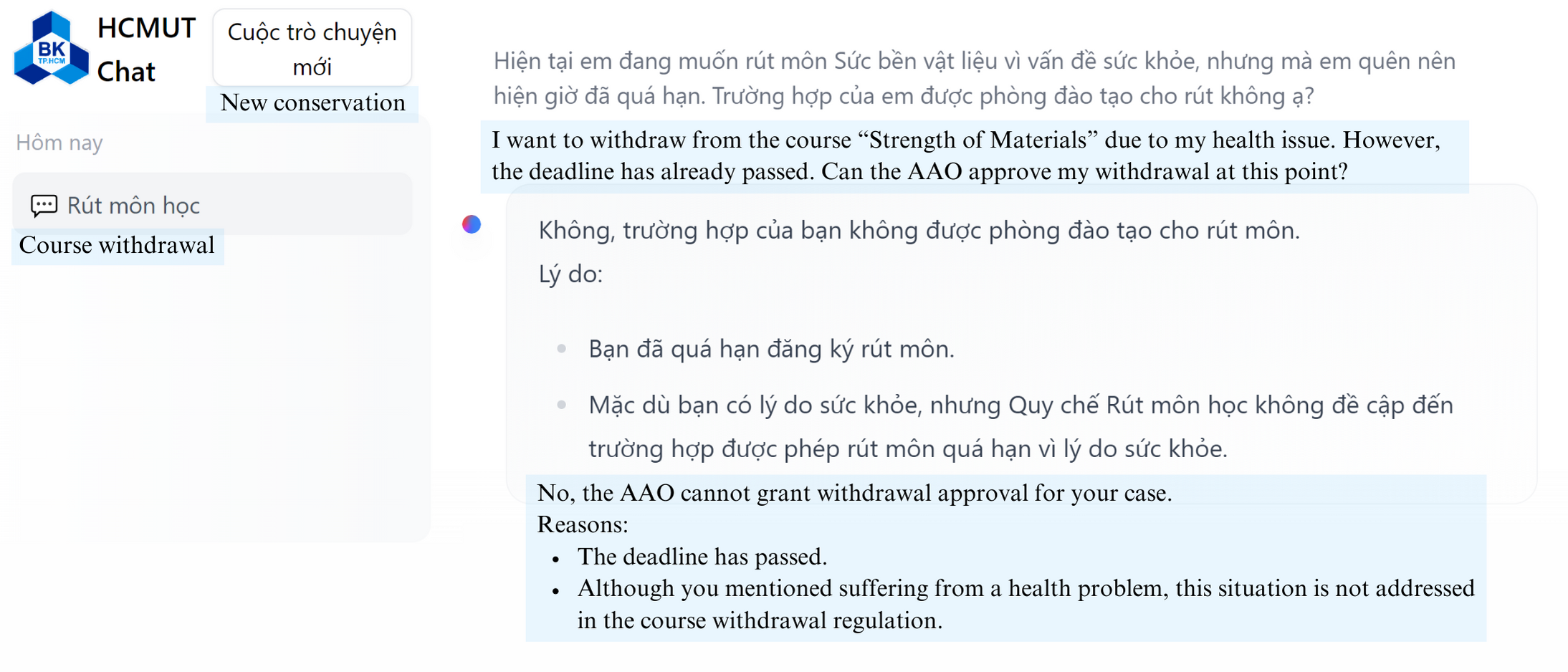}}
\vspace{-0.2cm}\caption{KG-augmented LLMs Approach and Demonstration of course withdrawal case}
\label{fig:figure_kgrag_pipeline_and_example}
\end{figure}

To address the "KG-augmented LLMs" aspect of the research and its application in the education domain at HCMUT, we
conduct a general pipeline of the question-answering system following the Figure \ref{fig:figure_kgrag_pipeline_and_example}  (a). The process of answering user
questions involves feeding the question into our institution’s LLM, URA, which generates a structured query in the
form of Neo4j’s Cypher language, as our knowledge graph database is built on Neo4j. This query searches a knowledge
graph database handcrafted using intent discovery’s result to retrieve matching sub-graphs and relevant triples. Next,
the triples are converted into a natural language format and combined with the original user question to create a final
prompt for the LLM. The prompt guides the LLM in generating a coherent and accurate answer as demonstrated in
Figure \ref{fig:figure_kgrag_pipeline_and_example}  (b).

\subsection{Discussion}

The section above outlines the outcomes of our research efforts focused on three key areas: Discovering open-intents, Developing an embedding-based technique for Relation Discovery, and Implementing KG-augmented LLMs approach.

First, discovering open-intents through unsupervised approach shows promising results, but applying it to Vietnamese encounters various difficulties due to the language's specific characteristics. There is a considerable amount of overlapping and repetitive clusters due to the inadequate performance of NLP toolkits for Vietnamese. Named Entity (NE) within sentences further interferes with the clustering process. Due to the subpar clustering step, automatic labeling is also affected accordingly. Applying Hyponym-approach to these NERs represents a potential approach for optimizing clustering. 

Second, the Embedding-based method for Relation Discovery has shown promising results in identifying relationships between entities. However, further research is needed to explore the different types of relations that can be discovered using this method. Additionally, there is a need to develop more robust and accurate methods for relation extraction and classification.

Finally, the KG-augmented LLMs approach is initially implemented by using off-the-shelf tools from Neo4J. The KG-retrieval engine extracts triplets. The advantages of this method are that the information is condensed and accurate, and the triplets also represent relations to help the LLM answer questions better. However, the triplets retrieval engine only extracts relevant triples and does not have a ranking mechanism to help eliminate redundant triples.

\section{Conclusion}

In this article, we introduce an approach to open intent discovery using unsupervised learning methods. To the best of our knowledge, this is the first paper to use this approach for open intent discovery in Vietnamese. The results are evaluated on three datasets: Bankinf77\_eng, Banking77\_vni, and HCMUT\_FAQ\_vni. Additionally, we perform preliminary experiments on discovering the relationships between intents and other entities to build a KG for the education domain. We also conduct some experiments on applying the KG to LLM.



\bibliographystyle{ACM-Reference-Format}
\bibliography{paper_01_ref}


\newpage
\appendix

\section{Appendix}
\subsection{E-OED Framework Analysis}
This appendix provides further details on the implementation of the E-OED framework. We discuss key design choices, such as hyperparameter selection and embedding models, along with the insights gained from related experiments.  

This section will also discuss three core modules in E-OED frameworks known as \textit{Data Preprocessing}, \textit{Semantic Clustering} and \textit{Automatic Cluster Labeling} 

The result achieved by different design choices is represented in detail at \ref{tab:design_selection_result}
\subsubsection{Data Preprocessing}
Due to Vietnamese being a predominantly analytic language with a large number of monosyllabic morphemes (meaningful units), words are often formed by combining these morphemes. This can present challenges for the original BERTopic framework, which relies on cTF-IDF (class-based TF-IDF)\cite{grootendorst2022bertopic} for topic modeling. To address this, we implemented an additional preprocessing step that leverages a word segmentation model to combine sub-word units into complete words. These newly formed words are then joined using an underscore character ("\_"). (e.g., subject-"môn học" becomes "môn\_học", course-"khóa học" becomes "khóa\_học").

Furthermore, to comply with dataset privacy policies, we implemented a text anonymization tool to censor any personally identifiable information (PII) within the sentences. Details regarding the specific objects censored and the corresponding replacement terms are presented in Table \ref{tab:data_censored}

\begin{table}[H]
  \caption{Data censored tool}
  \label{tab:data_censored}
  \begin{adjustbox}{width=0.4\textwidth}
  \begin{tabular}{lccc}
    \toprule
    \textbf{Objects}  & \textbf{Method} & \textbf{Term Replace}    \\
    \midrule
    Person Name & NER (BERT) & [full\_name]   \\
    Student ID & Regex (7 number format) & [student\_id]     \\
    Phone number & Regex (10 number format)   & [phone\_number]    \\
    Email & Regex (ordinary email with @) & [email]     \\ 
    \bottomrule
  \end{tabular}
  \end{adjustbox}
\end{table}

\subsubsection{Semantic Clustering}
Our model consists of three primary components:  
\begin{itemize}
    \item Sentence Embedding: extracts vector representations of sentences.
    \item Dimension Reduction: addresses the "curse of dimensionality" by compressing these embeddings into a lower-dimensional space, improving computational efficiency.
    \item Sentence-density clustering: groups sentences based on their reduced-dimensional representations. 
\end{itemize}
Our model's performance is highly sensitive to the chosen hyperparameters. We explored various options for each module:

Sentence Embedding:
\begin{itemize}
    \item \textit{BERTopic Base Model (MiniLM-L6)}: A 6-layer Microsoft MiniLM \cite{wang2020minilm} fine-tuned on a 1 billion sentence pair dataset.
    \item \textit{Supervised Vietnamese SimCSE}: Fine-tuned from the PhoBERT model.
\end{itemize}

Dimension Reduction (UMAP):
\begin{itemize}
    \item \textit{Number of neighbors}: 20, \textit{Number of components}: 4
    \item \textit{Number of neighbors}: 15, \textit{Number of components}: 9
\end{itemize}

Sentence-Density Clustering (HDBSCAN):
\begin{itemize}
    \item \textit{Minimum Cluster Size}: 15
    \item \textit{Minimum Cluster Size}: 20
\end{itemize}
We evaluated different combinations of these hyperparameters to identify the optimal configuration for our task.

\subsubsection{Automatic Cluster Labeling}
This section details the process for extracting descriptive labels for our identified clusters. We first select the seven most representative sentences for each cluster. These sentences are chosen based on their similarity to keywords generated using cTF-IDF.

Within this approach, a key hyperparameter is the range of word lengths considered for keyword extraction. We experiment with two settings: a range of 5-7 words and a range of 4-11 words. The first setting targets terms with lengths between 5 and 7 words, while the second allows for terms between 4 and 11 words.

Following keyword extraction, the chosen sentences are processed using the PhoNLP model to obtain Part-of-Speech (PoS) and dependency tags. We then leverage these tags in conjunction with pre-defined rules (detailed in the Algorithm \ref{algo:1}) to generate descriptive labels for each cluster.

\begin{algorithm}[H]
\scriptsize 
\caption{Tag extraction algorithms}
\label{algo:1}
\begin{algorithmic}[1]
\Function{ExtractSentenceElements}{$category\_docs$}
   \State Initialize list of labels
    \For{each sentence in $category\_docs$}
        \State Clean the sentence by stripping whitespace
        \State Get the annotation result from $PhoNLP$ 
        \State Extract the part-of-speech (POS) and dependency (DEP) information
        \State Initialize lists for verbs, direct objects, verb modifiers, and other list
        \State Initialize variables for root position, direct object positions, verb modifier position, and preposition position
        
        \textbf{\#Extract all related positions}
        \State Create a set to store the relevant positions
        \For{each token in the sentence}
            \If{the token is the root of the sentence}
                \State Add its position to the set
            \EndIf
            \While{the length of the position set is changing}
                \For{each token in the document}
                    \If{the token's dependency is related to the root or other relevant positions}
                        \State Add its position to the set
                    \EndIf
                \EndFor
                \If{the length of the set doesn't change}
                    \State Break the loop
                \EndIf
            \EndWhile
        \EndFor
        
        \textbf{\#Extract the sentence elements}
        \For{each token in the document}
            \If{the token's dependency is 'root'}
                \State Add it to the verbs and consecutive lists
            \EndIf
            \If{the token's dependency is 'prp' and its position is in the set}
                \State Add it to the direct objects lists
            \EndIf
            \If{the token's dependency is 'dob'}
                \State Add it to the direct objects list
            \EndIf
            \If{the token's dependency is 'vmod' and its position is related to the root, preposition, or verb modifier positions}
                \State Add it to the verb modifiers list
            \EndIf
            \If{the token's dependency is 'nmod' and its position is related to a direct object position}
                \State Add it to the other list
            \EndIf
            \If{the token's position or its dependency's position is in the set and the dependency is in a predefined list}
                \State Add the token to the other list
            \EndIf
        \EndFor
        \State Extract the most frequent tokens in each list.
        \State Add the tokens into list of labels
    \EndFor
    \State \textbf{return} list of labels
\EndFunction
\end{algorithmic}
\end{algorithm}

In order to reduce noises on newly extracted tags, we then take the most frequently occurring tokens within each subset, which are then designated as our cluster labels. 
For better cluster labels, LLMs can be used.

\subsubsection{Result and Insights}
There are 5 experiments with different design selection in total:
\begin{samepage}
\begin{enumerate}
  \item \textbf{Experiment 1:}
    \begin{itemize}
      \item Dataset: Banking77\_eng
      \item Number of Neighbors: 20
      \item Number of Components: 4
      \item Minimum Cluster Size: 20
      \item cTF-IDF Range: [5-7]
    \end{itemize}
  \item \textbf{Experiment 2:}
    \begin{itemize}
      \item Dataset: Banking77\_vni 
    \end{itemize}
    \begin{enumerate}
      \item \textbf{Experiment 2.1}:
        \begin{itemize}
          \item Number of Neighbors: 20
          \item Number of Components: 4
          \item Minimum Cluster Size: 20
          \item cTF-IDF Range: [5-7]
        \end{itemize}
      \item \textbf{Experiment 2.2}:
        \begin{itemize}
          \item Number of Neighbors: 15
          \item Number of Components: 9
          \item Minimum Cluster Size: 15
          \item cTF-IDF Range: [4-11]
        \end{itemize}
    \end{enumerate}
  \item \textbf{Experiment 3:}
    \begin{itemize}
      \item Dataset: Banking77\_vni
      \item Language Model: SimCSE (instead of MiniLM)
    \end{itemize}
    \begin{enumerate}
      \item \textbf{Experiment 3.1}
        \begin{itemize}
          \item Number of Neighbors: 20
          \item Number of Topics: 4
          \item Minimum Cluster Size: 20
          \item cTF-IDF Range: [5-7]
        \end{itemize}
      \item \textbf{Experiment 3.2 }
        \begin{itemize}
          \item Number of Neighbors: 15
          \item Number of Components: 9
          \item Minimum Cluster Size: 15
          \item cTF-IDF Range: [4-11]
        \end{itemize}
    \end{enumerate}
\end{enumerate}
\end{samepage}

\begin{table}[h]
\centering
  \caption{Design selection result}
  \label{tab:design_selection_result}
   \begin{adjustbox}{width=0.25\textwidth}
  \begin{tabular}{lccc}
    \toprule
    \textbf{Experiment}  & \textbf{Intentions Discovered}\\
    \midrule
    Experiment 1 & 73  \\
    Experiment 2.1 & 65 \\
    Experiment 2.2 & 74 \\
    Experiment 3.1 & 71 \\ 
    Experiment 3.2 & \textbf{76} \\ 
    \bottomrule
  \end{tabular}
   \end{adjustbox}
\end{table}

Our experiments revealed several key insights. Firstly, the original embedding model, despite not being trained on Vietnamese data, achieved reasonable performance on our banking77\_vni dataset. This suggests potential transferability due to structural similarities across languages. However, fine-tuning a dedicated embedding model specifically for Vietnamese could likely yield further improvements. Evaluating the quality of the embedding model can be achieved by analyzing the representative sentences within each cluster. Greater semantic similarity within clusters signifies a more effective model.

Secondly, a higher number of components in UMAP increases focus on local attributes, leading to greater separation and potentially clearer visualizations. However, it's crucial to be mindful of the "curse of dimensionality" when employing this approach.

Finally, the framework is sensitive to hyperparameter selection. Due to the presence of numerous hyperparameters requiring tuning, careful consideration should be given when applying this method to your specific domain or FAQ dataset.
\end{document}

%% file: latex_figures/figure_01_multisource_data.tex
\tikzset{every picture/.style={line width=0.75pt}}
\hspace*{-2.6cm}\begin{tikzpicture}[scale=0.5,x=0.75pt,y=0.75pt,yscale=-1,xscale=1]

\draw  [fill={rgb, 255:red, 240; green, 240; blue, 240 }  ,fill opacity=1 ] (20,44.8) .. controls (20,34.97) and (27.97,27) .. (37.8,27) -- (642.2,27) .. controls (652.03,27) and (660,34.97) .. (660,44.8) -- (660,98.2) .. controls (660,108.03) and (652.03,116) .. (642.2,116) -- (37.8,116) .. controls (27.97,116) and (20,108.03) .. (20,98.2) -- cycle ;
\draw  [fill={rgb, 255:red, 240; green, 240; blue, 240 }  ,fill opacity=1 ] (20,163.8) .. controls (20,153.97) and (27.97,146) .. (37.8,146) -- (642.2,146) .. controls (652.03,146) and (660,153.97) .. (660,163.8) -- (660,217.2) .. controls (660,227.03) and (652.03,235) .. (642.2,235) -- (37.8,235) .. controls (27.97,235) and (20,227.03) .. (20,217.2) -- cycle ;
\draw  [fill={rgb, 255:red, 255; green, 255; blue, 255 }  ,fill opacity=1 ] (46.1,169.6) .. controls (46.1,164.57) and (50.17,160.5) .. (55.2,160.5) -- (178.8,160.5) .. controls (183.83,160.5) and (187.9,164.57) .. (187.9,169.6) -- (187.9,196.9) .. controls (187.9,201.93) and (183.83,206) .. (178.8,206) -- (55.2,206) .. controls (50.17,206) and (46.1,201.93) .. (46.1,196.9) -- cycle ;
\draw [color={rgb, 255:red, 128; green, 128; blue, 128 }  ,draw opacity=1 ]   (115,100) -- (115,160) ;
\draw [color={rgb, 255:red, 128; green, 128; blue, 128 }  ,draw opacity=1 ]   (140,100) -- (320,190) ;
\draw [color={rgb, 255:red, 128; green, 128; blue, 128 }  ,draw opacity=1 ]   (282,100) -- (310,180) ;
\draw [color={rgb, 255:red, 128; green, 128; blue, 128 }  ,draw opacity=1 ]   (440,90) -- (380,180) ;
\draw [color={rgb, 255:red, 128; green, 128; blue, 128 }  ,draw opacity=1 ]   (570,90) -- (380,190) ;
\draw [color={rgb, 255:red, 128; green, 128; blue, 128 }  ,draw opacity=1 ]   (573,100) -- (573,160) ;
\draw  [fill={rgb, 255:red, 255; green, 255; blue, 255 }  ,fill opacity=1 ] (275.2,183.25) .. controls (275.2,170.69) and (307.39,160.5) .. (347.1,160.5) .. controls (386.81,160.5) and (419,170.69) .. (419,183.25) .. controls (419,195.81) and (386.81,206) .. (347.1,206) .. controls (307.39,206) and (275.2,195.81) .. (275.2,183.25) -- cycle ;
\draw   (321.57,264.92) -- (328.54,264.92) -- (328.54,242.3) -- (342.46,242.3) -- (342.46,264.92) -- (349.43,264.92) -- (335.5,280) -- cycle ;
\draw   (60.5,287) -- (620,287) -- (620,340) -- (60.5,340) -- cycle(612.05,294.95) -- (68.45,294.95) -- (68.45,332.05) -- (612.05,332.05) -- cycle ;
\draw  [fill={rgb, 255:red, 255; green, 255; blue, 255 }  ,fill opacity=1 ] (511.88,160) -- (647,160) -- (632.12,205.5) -- (497,205.5) -- cycle ;

\draw  [fill={rgb, 255:red, 255; green, 255; blue, 255 }  ,fill opacity=1 ]  (33,55.25) -- (200,55.25) -- (200,104.25) -- (33,104.25) -- cycle  ;
\draw (116.5,79.75) node   [align=left] {\begin{minipage}[lt]{110.84pt}\setlength\topsep{0pt}
\begin{center} \tiny
{\tiny\textbf{FAQ \& HelpDesk System}}
\end{center}

\end{minipage}};
\draw  [fill={rgb, 255:red, 255; green, 255; blue, 255 }  ,fill opacity=1 ]  (215,55.25) -- (344,55.25) -- (344,104.25) -- (215,104.25) -- cycle  ;
\draw (279.5,79.75) node   [align=left] {\begin{minipage}[lt]{85pt}\setlength\topsep{0pt}
\begin{center} \tiny
\textbf{Portals / Websites}
\end{center}

\end{minipage}};
\draw  [fill={rgb, 255:red, 255; green, 255; blue, 255 }  ,fill opacity=1 ]  (360,55.25) -- (489,55.25) -- (489,104.25) -- (360,104.25) -- cycle  ;
\draw (424.5,79.75) node   [align=left] {\begin{minipage}[lt]{85pt}\setlength\topsep{0pt}
\begin{center} \tiny
\textbf{LMS}
\end{center}

\end{minipage}};
\draw  [fill={rgb, 255:red, 255; green, 255; blue, 255 }  ,fill opacity=1 ]  (505,55.25) -- (648,55.25) -- (648,104.25) -- (505,104.25) -- cycle  ;
\draw (576.5,79.75) node   [align=left] {\begin{minipage}[lt]{94.52pt}\setlength\topsep{0pt}
\begin{center} \tiny
\textbf{Other systems}
\end{center}

\end{minipage}};
\draw (156,42) node   [align=left] {\begin{minipage}[lt]{92.21pt}\setlength\topsep{0pt}
{\tiny\textit{University Resources}}
\end{minipage}};
\draw (117,183.25) node   [align=left] {\begin{minipage}[lt]{96.42pt}\setlength\topsep{0pt}
\begin{center} \tiny
{\tiny\textbf{Intention entity}}
\end{center}

\end{minipage}};
\draw (347.1,183.25) node   [align=left] {\begin{minipage}[lt]{97.78pt}\setlength\topsep{0pt}
\begin{center} \tiny
\textbf{Academic entity}
\end{center}

\end{minipage}};
\draw (235,221) node  [font=\tiny] [align=left] {\begin{minipage}[lt]{150pt}\setlength\topsep{0pt}
\textit{Educational Open Entity Discovery}
\end{minipage}};
\draw (340.25,313.5) node   [align=left] {\begin{minipage}[lt]{380.46pt}\setlength\topsep{0pt}
\begin{center} \tiny
\textbf{HCMUT LLM-based Virtual Assistant}
\end{center}

\end{minipage}};
\draw (572,182.75) node   [align=left] {\begin{minipage}[lt]{102pt}\setlength\topsep{0pt}
\begin{center} \tiny
\textbf{Policy / Document}
\end{center}

\end{minipage}};

\end{tikzpicture}

%% file: latex_figures/figure_02_education_knowledge_graph.tex
\tikzset{every picture/.style={line width=0.75pt}}
\hspace*{-1.65cm}\begin{tikzpicture}[scale=0.179,x=1.75pt,y=1.6pt,yscale=-1,xscale=1]

\draw [color={rgb, 255:red, 128; green, 128; blue, 128 }  ,draw opacity=1 ]   (712.99,1378.78) -- (716,1481.06) ;
\draw [color={rgb, 255:red, 128; green, 128; blue, 128 }  ,draw opacity=1 ]   (612,1197.04) -- (584.57,1308.21) ;
\draw [color={rgb, 255:red, 128; green, 128; blue, 128 }  ,draw opacity=1 ]   (313,1275.56) .. controls (-70,1157.04) and (7,1659.06) .. (377.87,1588.28) ;
\draw [color={rgb, 255:red, 128; green, 128; blue, 128 }  ,draw opacity=1 ]   (406,1507.06) -- (469,1448.54) ;
\draw [color={rgb, 255:red, 128; green, 128; blue, 128 }  ,draw opacity=1 ]   (455.87,1585.28) .. controls (704,1626.06) and (772,1512.06) .. (779,1481.06) ;
\draw [color={rgb, 255:red, 128; green, 128; blue, 128 }  ,draw opacity=1 ]   (585.14,1342.04) .. controls (523.4,1427.29) and (359.22,1360.41) .. (246,1415.06) ;
\draw [color={rgb, 255:red, 128; green, 128; blue, 128 }  ,draw opacity=1 ]   (188,1212.04) -- (277,1166.04) ;
\draw [color={rgb, 255:red, 128; green, 128; blue, 128 }  ,draw opacity=1 ]   (302,1432.06) -- (215,1481.06) ;
\draw [color={rgb, 255:red, 128; green, 128; blue, 128 }  ,draw opacity=1 ]   (637,1519.54) -- (587,1452.54) ;
\draw [color={rgb, 255:red, 128; green, 128; blue, 128 }  ,draw opacity=1 ]   (744,1349.56) -- (836,1402.06) ;
\draw [color={rgb, 255:red, 128; green, 128; blue, 128 }  ,draw opacity=1 ]   (613,1326.04) -- (785,1306.04) ;
\draw [color={rgb, 255:red, 128; green, 128; blue, 128 }  ,draw opacity=1 ]   (269,1322.54) -- (225,1307.54) ;
\draw [color={rgb, 255:red, 128; green, 128; blue, 128 }  ,draw opacity=1 ]   (192,1394.06) -- (262.82,1343.53) ;
\draw [color={rgb, 255:red, 128; green, 128; blue, 128 }  ,draw opacity=1 ]   (371,1283.56) .. controls (490,1345.56) and (499,1279.56) .. (758,1291.56) ;
\draw [color={rgb, 255:red, 128; green, 128; blue, 128 }  ,draw opacity=1 ]   (374,1289.56) .. controls (381,1430.56) and (509,1411.04) .. (507,1452.54) ;
\draw [color={rgb, 255:red, 128; green, 128; blue, 128 }  ,draw opacity=1 ]   (358.11,1414.8) -- (361.61,1293.68) ;
\draw [color={rgb, 255:red, 128; green, 128; blue, 128 }  ,draw opacity=1 ]   (347,1295.56) .. controls (351,1402.06) and (274.01,1367.01) .. (246,1398.06) ;
\draw [color={rgb, 255:red, 128; green, 128; blue, 128 }  ,draw opacity=1 ]   (377.37,1276.6) .. controls (422.51,1231.83) and (516.15,1240.27) .. (590.97,1244.06) .. controls (629.9,1246.03) and (663.73,1246.73) .. (683,1238.04) ;
\draw [color={rgb, 255:red, 128; green, 128; blue, 128 }  ,draw opacity=1 ]   (362,1260.06) .. controls (438.04,1114.88) and (614.69,1194.45) .. (671,1169.04) ;
\draw [color={rgb, 255:red, 128; green, 128; blue, 128 }  ,draw opacity=1 ]   (672,1252.04) -- (597.97,1315.11) ;
\draw [color={rgb, 255:red, 128; green, 128; blue, 128 }  ,draw opacity=1 ]   (359,1271.54) -- (457,1214.54) ;
\draw [color={rgb, 255:red, 128; green, 128; blue, 128 }  ,draw opacity=1 ]   (330,1266.04) -- (225.85,1213.03) ;
\draw [color={rgb, 255:red, 128; green, 128; blue, 128 }  ,draw opacity=1 ]   (398.67,1165.56) -- (457.67,1194.9) ;
\draw [color={rgb, 255:red, 128; green, 128; blue, 128 }  ,draw opacity=1 ]   (488.54,1227.54) -- (460.53,1262.94) ;
\draw [color={rgb, 255:red, 128; green, 128; blue, 128 }  ,draw opacity=1 ]   (510.67,1221.56) -- (569.33,1317.56) ;
\draw [color={rgb, 255:red, 128; green, 128; blue, 128 }  ,draw opacity=1 ]   (513,1432.54) -- (601,1339.56) ;
\draw [color={rgb, 255:red, 128; green, 128; blue, 128 }  ,draw opacity=1 ]   (697,1363.06) -- (619.99,1334.83) ;
\draw [color={rgb, 255:red, 128; green, 128; blue, 128 }  ,draw opacity=1 ]   (659,1403.54) -- (600,1334.54) ;
\draw [color={rgb, 255:red, 128; green, 128; blue, 128 }  ,draw opacity=1 ]   (386.12,1421.32) -- (577.84,1343.99) ;
\draw [color={rgb, 255:red, 128; green, 128; blue, 128 }  ,draw opacity=1 ]   (478.91,1352.38) -- (544.57,1333.74) ;
\draw [color={rgb, 255:red, 128; green, 128; blue, 128 }  ,draw opacity=1 ]   (487.67,1286.54) -- (548.07,1320.08) ;
\draw  [fill={rgb, 255:red, 255; green, 255; blue, 255 }  ,fill opacity=1 ][blur shadow={shadow xshift=0.5pt,shadow yshift=-0.5pt, shadow blur radius=2pt, shadow blur steps=4 ,shadow opacity=50}] (409.56,1276.58) .. controls (409.56,1264.61) and (430.15,1254.91) .. (455.56,1254.91) .. controls (480.96,1254.91) and (501.56,1264.61) .. (501.56,1276.58) .. controls (501.56,1288.54) and (480.96,1298.24) .. (455.56,1298.24) .. controls (430.15,1298.24) and (409.56,1288.54) .. (409.56,1276.58) -- cycle ;
\draw  [fill={rgb, 255:red, 255; green, 255; blue, 255 }  ,fill opacity=1 ][blur shadow={shadow xshift=0.5pt,shadow yshift=-0.5pt, shadow blur radius=2pt, shadow blur steps=4 ,shadow opacity=50}] (538.57,1329.88) .. controls (538.57,1317.91) and (559.16,1308.21) .. (584.57,1308.21) .. controls (609.97,1308.21) and (630.57,1317.91) .. (630.57,1329.88) .. controls (630.57,1341.84) and (609.97,1351.55) .. (584.57,1351.55) .. controls (559.16,1351.55) and (538.57,1341.84) .. (538.57,1329.88) -- cycle ;
\draw  [fill={rgb, 255:red, 255; green, 255; blue, 255 }  ,fill opacity=1 ][blur shadow={shadow xshift=0.5pt,shadow yshift=-0.5pt, shadow blur radius=2pt, shadow blur steps=4 ,shadow opacity=50}] (611.47,1417.64) .. controls (611.47,1405.67) and (632.07,1395.97) .. (657.47,1395.97) .. controls (682.88,1395.97) and (703.47,1405.67) .. (703.47,1417.64) .. controls (703.47,1429.61) and (682.88,1439.31) .. (657.47,1439.31) .. controls (632.07,1439.31) and (611.47,1429.61) .. (611.47,1417.64) -- cycle ;
\draw  [fill={rgb, 255:red, 255; green, 255; blue, 255 }  ,fill opacity=1 ][blur shadow={shadow xshift=0.5pt,shadow yshift=-0.5pt, shadow blur radius=2pt, shadow blur steps=4 ,shadow opacity=50}] (666.99,1357.11) .. controls (666.99,1345.15) and (687.59,1335.44) .. (712.99,1335.44) .. controls (738.4,1335.44) and (758.99,1345.15) .. (758.99,1357.11) .. controls (758.99,1369.08) and (738.4,1378.78) .. (712.99,1378.78) .. controls (687.59,1378.78) and (666.99,1369.08) .. (666.99,1357.11) -- cycle ;
\draw  [fill={rgb, 255:red, 255; green, 255; blue, 255 }  ,fill opacity=1 ][blur shadow={shadow xshift=0.5pt,shadow yshift=-0.5pt, shadow blur radius=2pt, shadow blur steps=4 ,shadow opacity=50}] (396.77,1348.6) .. controls (396.77,1336.63) and (417.37,1326.93) .. (442.77,1326.93) .. controls (468.18,1326.93) and (488.77,1336.63) .. (488.77,1348.6) .. controls (488.77,1360.57) and (468.18,1370.27) .. (442.77,1370.27) .. controls (417.37,1370.27) and (396.77,1360.57) .. (396.77,1348.6) -- cycle ;
\draw  [fill={rgb, 255:red, 240; green, 240; blue, 240 }  ,fill opacity=1 ][blur shadow={shadow xshift=0.5pt,shadow yshift=-0.5pt, shadow blur radius=2pt, shadow blur steps=4 ,shadow opacity=50}] (446.34,1424.94) .. controls (446.34,1420.15) and (450.22,1416.27) .. (455.01,1416.27) -- (594.06,1416.27) .. controls (598.84,1416.27) and (602.72,1420.15) .. (602.72,1424.94) -- (602.72,1450.94) .. controls (602.72,1455.73) and (598.84,1459.61) .. (594.06,1459.61) -- (455.01,1459.61) .. controls (450.22,1459.61) and (446.34,1455.73) .. (446.34,1450.94) -- cycle ;
\draw  [fill={rgb, 255:red, 240; green, 240; blue, 240 }  ,fill opacity=1 ][blur shadow={shadow xshift=0.5pt,shadow yshift=-0.5pt, shadow blur radius=2pt, shadow blur steps=4 ,shadow opacity=50}] (647.92,1232.91) .. controls (647.92,1228.13) and (651.8,1224.25) .. (656.59,1224.25) -- (775.33,1224.25) .. controls (780.12,1224.25) and (784,1228.13) .. (784,1232.91) -- (784,1258.91) .. controls (784,1263.7) and (780.12,1267.58) .. (775.33,1267.58) -- (656.59,1267.58) .. controls (651.8,1267.58) and (647.92,1263.7) .. (647.92,1258.91) -- cycle ;
\draw  [fill={rgb, 255:red, 240; green, 240; blue, 240 }  ,fill opacity=1 ][blur shadow={shadow xshift=0.5pt,shadow yshift=-0.5pt, shadow blur radius=2pt, shadow blur steps=4 ,shadow opacity=50}] (277,1420.32) .. controls (277,1415.53) and (280.88,1411.65) .. (285.67,1411.65) -- (425.33,1411.65) .. controls (430.12,1411.65) and (434,1415.53) .. (434,1420.32) -- (434,1446.32) .. controls (434,1451.1) and (430.12,1454.99) .. (425.33,1454.99) -- (285.67,1454.99) .. controls (280.88,1454.99) and (277,1451.1) .. (277,1446.32) -- cycle ;
\draw  [fill={rgb, 255:red, 255; green, 255; blue, 255 }  ,fill opacity=1 ][blur shadow={shadow xshift=0.5pt,shadow yshift=-0.5pt, shadow blur radius=2pt, shadow blur steps=4 ,shadow opacity=50}] (295.96,1280.91) .. controls (295.96,1268.94) and (316.55,1259.24) .. (341.96,1259.24) .. controls (367.36,1259.24) and (387.96,1268.94) .. (387.96,1280.91) .. controls (387.96,1292.88) and (367.36,1302.58) .. (341.96,1302.58) .. controls (316.55,1302.58) and (295.96,1292.88) .. (295.96,1280.91) -- cycle ;
\draw  [fill={rgb, 255:red, 255; green, 255; blue, 255 }  ,fill opacity=1 ][blur shadow={shadow xshift=0.5pt,shadow yshift=-0.5pt, shadow blur radius=2pt, shadow blur steps=4 ,shadow opacity=50}] (234.28,1336.21) .. controls (234.28,1324.24) and (257.05,1314.54) .. (285.14,1314.54) .. controls (313.23,1314.54) and (336,1324.24) .. (336,1336.21) .. controls (336,1348.17) and (313.23,1357.87) .. (285.14,1357.87) .. controls (257.05,1357.87) and (234.28,1348.17) .. (234.28,1336.21) -- cycle ;
\draw  [fill={rgb, 255:red, 255; green, 255; blue, 255 }  ,fill opacity=1 ][blur shadow={shadow xshift=0.5pt,shadow yshift=-0.5pt, shadow blur radius=2pt, shadow blur steps=4 ,shadow opacity=50}] (158,1220.98) .. controls (158,1207.13) and (185.76,1195.9) .. (220,1195.9) .. controls (254.24,1195.9) and (282,1207.13) .. (282,1220.98) .. controls (282,1234.83) and (254.24,1246.06) .. (220,1246.06) .. controls (185.76,1246.06) and (158,1234.83) .. (158,1220.98) -- cycle ;
\draw  [fill={rgb, 255:red, 240; green, 240; blue, 240 }  ,fill opacity=1 ][blur shadow={shadow xshift=0.5pt,shadow yshift=-0.5pt, shadow blur radius=2pt, shadow blur steps=4 ,shadow opacity=50}] (448.95,1193.71) .. controls (448.95,1188.92) and (452.83,1185.04) .. (457.61,1185.04) -- (560.8,1185.04) .. controls (565.59,1185.04) and (569.47,1188.92) .. (569.47,1193.71) -- (569.47,1219.71) .. controls (569.47,1224.49) and (565.59,1228.37) .. (560.8,1228.37) -- (457.61,1228.37) .. controls (452.83,1228.37) and (448.95,1224.49) .. (448.95,1219.71) -- cycle ;
\draw  [fill={rgb, 255:red, 255; green, 255; blue, 255 }  ,fill opacity=1 ][blur shadow={shadow xshift=0.5pt,shadow yshift=-0.5pt, shadow blur radius=2pt, shadow blur steps=4 ,shadow opacity=50}] (255.33,1177.48) .. controls (255.33,1161.88) and (292.72,1149.23) .. (338.83,1149.23) .. controls (384.95,1149.23) and (422.33,1161.88) .. (422.33,1177.48) .. controls (422.33,1193.08) and (384.95,1205.73) .. (338.83,1205.73) .. controls (292.72,1205.73) and (255.33,1193.08) .. (255.33,1177.48) -- cycle ;
\draw  [fill={rgb, 255:red, 240; green, 240; blue, 240 }  ,fill opacity=1 ][blur shadow={shadow xshift=0.5pt,shadow yshift=-0.5pt, shadow blur radius=2pt, shadow blur steps=4 ,shadow opacity=50}] (589.02,1172.83) .. controls (589.02,1168.04) and (592.9,1164.16) .. (597.68,1164.16) -- (729.32,1164.16) .. controls (734.11,1164.16) and (737.99,1168.04) .. (737.99,1172.83) -- (737.99,1198.83) .. controls (737.99,1203.62) and (734.11,1207.5) .. (729.32,1207.5) -- (597.68,1207.5) .. controls (592.9,1207.5) and (589.02,1203.62) .. (589.02,1198.83) -- cycle ;
\draw  [fill={rgb, 255:red, 240; green, 240; blue, 240 }  ,fill opacity=1 ][blur shadow={shadow xshift=0.5pt,shadow yshift=-0.5pt, shadow blur radius=2pt, shadow blur steps=4 ,shadow opacity=50}] (725.76,1290.84) .. controls (725.76,1286.06) and (729.64,1282.18) .. (734.42,1282.18) -- (855.33,1282.18) .. controls (860.12,1282.18) and (864,1286.06) .. (864,1290.84) -- (864,1316.84) .. controls (864,1321.63) and (860.12,1325.51) .. (855.33,1325.51) -- (734.42,1325.51) .. controls (729.64,1325.51) and (725.76,1321.63) .. (725.76,1316.84) -- cycle ;
\draw  [fill={rgb, 255:red, 255; green, 255; blue, 255 }  ,fill opacity=1 ][blur shadow={shadow xshift=0.5pt,shadow yshift=-0.5pt, shadow blur radius=2pt, shadow blur steps=4 ,shadow opacity=50}] (323,1518.44) .. controls (323,1502.83) and (368.67,1490.19) .. (425,1490.19) .. controls (481.33,1490.19) and (527,1502.83) .. (527,1518.44) .. controls (527,1534.04) and (481.33,1546.69) .. (425,1546.69) .. controls (368.67,1546.69) and (323,1534.04) .. (323,1518.44) -- cycle ;
\draw  [fill={rgb, 255:red, 255; green, 255; blue, 255 }  ,fill opacity=1 ][blur shadow={shadow xshift=0.5pt,shadow yshift=-0.5pt, shadow blur radius=2pt, shadow blur steps=4 ,shadow opacity=50}] (150.94,1470.31) -- (332,1470.31) -- (312.06,1522.31) -- (131,1522.31) -- cycle ;
\draw  [fill={rgb, 255:red, 255; green, 255; blue, 255 }  ,fill opacity=1 ][blur shadow={shadow xshift=0.5pt,shadow yshift=-0.5pt, shadow blur radius=2pt, shadow blur steps=4 ,shadow opacity=50}] (118.27,1284.13) -- (238.75,1284.13) -- (225.49,1336.14) -- (105,1336.14) -- cycle ;
\draw  [fill={rgb, 255:red, 255; green, 255; blue, 255 }  ,fill opacity=1 ][blur shadow={shadow xshift=0.5pt,shadow yshift=-0.5pt, shadow blur radius=2pt, shadow blur steps=4 ,shadow opacity=50}] (738.77,1383.2) -- (882,1383.2) -- (866.23,1435.21) -- (723,1435.21) -- cycle ;
\draw  [fill={rgb, 255:red, 255; green, 255; blue, 255 }  ,fill opacity=1 ][blur shadow={shadow xshift=0.5pt,shadow yshift=-0.5pt, shadow blur radius=2pt, shadow blur steps=4 ,shadow opacity=50}] (714.74,1476.08) .. controls (714.74,1463.06) and (738.51,1452.51) .. (767.83,1452.51) .. controls (797.15,1452.51) and (820.91,1463.06) .. (820.91,1476.08) .. controls (820.91,1489.1) and (797.15,1499.66) .. (767.83,1499.66) .. controls (738.51,1499.66) and (714.74,1489.1) .. (714.74,1476.08) -- cycle ;
\draw  [fill={rgb, 255:red, 255; green, 255; blue, 255 }  ,fill opacity=1 ][blur shadow={shadow xshift=0.5pt,shadow yshift=-0.5pt, shadow blur radius=2pt, shadow blur steps=4 ,shadow opacity=50}] (543.03,1501.18) -- (725,1501.18) -- (704.97,1553.18) -- (523,1553.18) -- cycle ;
\draw  [fill={rgb, 255:red, 255; green, 255; blue, 255 }  ,fill opacity=1 ][blur shadow={shadow xshift=0.5pt,shadow yshift=-0.5pt, shadow blur radius=2pt, shadow blur steps=4 ,shadow opacity=50}] (377.87,1588.28) .. controls (377.87,1576.31) and (398.46,1566.61) .. (423.87,1566.61) .. controls (449.27,1566.61) and (469.87,1576.31) .. (469.87,1588.28) .. controls (469.87,1600.24) and (449.27,1609.94) .. (423.87,1609.94) .. controls (398.46,1609.94) and (377.87,1600.24) .. (377.87,1588.28) -- cycle ;
\draw  [fill={rgb, 255:red, 240; green, 240; blue, 240 }  ,fill opacity=1 ][blur shadow={shadow xshift=0.5pt,shadow yshift=-0.5pt, shadow blur radius=2pt, shadow blur steps=4 ,shadow opacity=50}] (138,1395.23) .. controls (138,1390.44) and (141.88,1386.56) .. (146.67,1386.56) -- (246.11,1386.56) .. controls (250.9,1386.56) and (254.78,1390.44) .. (254.78,1395.23) -- (254.78,1421.23) .. controls (254.78,1426.02) and (250.9,1429.9) .. (246.11,1429.9) -- (146.67,1429.9) .. controls (141.88,1429.9) and (138,1426.02) .. (138,1421.23) -- cycle ;
\draw [color={rgb, 255:red, 128; green, 128; blue, 128 }  ,draw opacity=1 ]   (409.56,1276.58) -- (387.96,1280.91) ;

\draw (455.56,1276.58) node   [align=left] {\begin{minipage}[lt]{62.56pt}\setlength\topsep{0pt}
\begin{center} \tiny
Class
\end{center}

\end{minipage}};
\draw (584.57,1329.88) node   [align=left] {\begin{minipage}[lt]{62.56pt}\setlength\topsep{0pt}
\begin{center} \tiny
Course
\end{center}

\end{minipage}};
\draw (657.47,1417.64) node   [align=left] {\begin{minipage}[lt]{62.56pt}\setlength\topsep{0pt}
\begin{center} \tiny
Semester
\end{center}

\end{minipage}};
\draw (712.99,1357.11) node   [align=left] {\begin{minipage}[lt]{62.56pt}\setlength\topsep{0pt}
\begin{center} \tiny
Subject
\end{center}

\end{minipage}};
\draw (442.77,1348.6) node   [align=left] {\begin{minipage}[lt]{62.56pt}\setlength\topsep{0pt}
\begin{center} \tiny
Lecturer
\end{center}

\end{minipage}};
\draw (524.53,1437.94) node   [align=left] {\begin{minipage}[lt]{106.36pt}\setlength\topsep{0pt}
\begin{center} \tiny
Course registration
\end{center}

\end{minipage}};
\draw (715.96,1245.91) node   [align=left] {\begin{minipage}[lt]{92.55pt}\setlength\topsep{0pt}
\begin{center} \tiny
Course deferral
\end{center}

\end{minipage}};
\draw (355.5,1433.32) node   [align=left] {\begin{minipage}[lt]{106.76pt}\setlength\topsep{0pt}
\begin{center} \tiny
Course withdrawal
\end{center}

\end{minipage}};
\draw (341.96,1280.91) node   [align=left] {\begin{minipage}[lt]{62.56pt}\setlength\topsep{0pt}
\begin{center} \tiny
Student
\end{center}

\end{minipage}};
\draw (285.14,1336.21) node   [align=left] {\begin{minipage}[lt]{69.16pt}\setlength\topsep{0pt}
\begin{center} \tiny
Tuition fee
\end{center}

\end{minipage}};
\draw (220,1220.98) node   [align=left] {\begin{minipage}[lt]{84.32pt}\setlength\topsep{0pt}
\begin{center} \tiny
Student status
\end{center}

\end{minipage}};
\draw (509.21,1206.71) node   [align=left] {\begin{minipage}[lt]{81.94pt}\setlength\topsep{0pt}
\begin{center} \tiny
Class transfer
\end{center}

\end{minipage}};
\draw (338.83,1177.48) node   [align=left] {\begin{minipage}[lt]{114.93pt}\setlength\topsep{0pt}
\begin{center} \tiny
Academic affairs office
\end{center}

\end{minipage}};
\draw (663.5,1185.83) node   [align=left] {\begin{minipage}[lt]{101.33pt}\setlength\topsep{0pt}
\begin{center} \tiny
Transcript inquiry
\end{center}

\end{minipage}};
\draw (794.88,1303.84) node   [align=left] {\begin{minipage}[lt]{93.98pt}\setlength\topsep{0pt}
\begin{center} \tiny
Grade complaint
\end{center}

\end{minipage}};
\draw (425,1518.44) node   [align=left] {\begin{minipage}[lt]{138.72pt}\setlength\topsep{0pt}
\begin{center} \tiny
Course registration schedule
\end{center}

\end{minipage}};
\draw (231.5,1496.31) node   [align=left] {\begin{minipage}[lt]{136.68pt}\setlength\topsep{0pt}
\begin{center} \tiny
Course withdrawal policy
\end{center}

\end{minipage}};
\draw (171.14,1310.1) node   [align=left] {\begin{minipage}[lt]{89.95pt}\setlength\topsep{0pt}
\begin{center} \tiny
Tuition policy
\end{center}

\end{minipage}};
\draw (802.5,1409.21) node   [align=left] {\begin{minipage}[lt]{108.12pt}\setlength\topsep{0pt}
\begin{center} \tiny
Subject information
\end{center}

\end{minipage}};
\draw (767.83,1476.08) node   [align=left] {\begin{minipage}[lt]{71.52pt}\setlength\topsep{0pt}
\begin{center} \tiny
Department
\end{center}

\end{minipage}};
\draw (624,1527.18) node   [align=left] {\begin{minipage}[lt]{137.36pt}\setlength\topsep{0pt}
\begin{center} \tiny
Course registration policy
\end{center}

\end{minipage}};
\draw (423.87,1588.28) node   [align=left] {\begin{minipage}[lt]{62.56pt}\setlength\topsep{0pt}
\begin{center} \tiny
Faculty
\end{center}

\end{minipage}};
\draw (196.39,1408.23) node   [align=left] {\begin{minipage}[lt]{78.42pt}\setlength\topsep{0pt}
\begin{center} \tiny
Course drop
\end{center}

\end{minipage}};

\end{tikzpicture}

%% file: latex_figures/figure_03_intent_discovery_pipeline.tex
\tikzset{every picture/.style={line width=0.75pt}} 
\hspace*{-0.3cm}\begin{tikzpicture}[scale=0.36,x=0.95pt,y=0.8pt,yscale=-1,xscale=1]

\draw   (454.67,1693.04) -- (677,1693.04) -- (677,1837.27) -- (454.67,1837.27) -- cycle ;
\draw [color={rgb, 255:red, 128; green, 128; blue, 128 }  ,draw opacity=1 ][fill={rgb, 255:red, 155; green, 155; blue, 155 }  ,fill opacity=1 ]   (335,1728.27) -- (406,1728.27) ;
\draw [shift={(409,1728.27)}, rotate = 180] [fill={rgb, 255:red, 128; green, 128; blue, 128 }  ,fill opacity=1 ][line width=0.08]  [draw opacity=0] (7.14,-3.43) -- (0,0) -- (7.14,3.43) -- (4.74,0) -- cycle    ;
\draw   (424.84,1862.54) -- (707.5,1862.54) -- (707.5,2007.27) -- (424.84,2007.27) -- cycle ;
\draw [color={rgb, 255:red, 128; green, 128; blue, 128 }  ,draw opacity=1 ][fill={rgb, 255:red, 155; green, 155; blue, 155 }  ,fill opacity=1 ]   (565.67,1837.27) -- (565.67,1859.06) ;
\draw [shift={(565.67,1862.06)}, rotate = 270] [fill={rgb, 255:red, 128; green, 128; blue, 128 }  ,fill opacity=1 ][line width=0.08]  [draw opacity=0] (7.14,-3.43) -- (0,0) -- (7.14,3.43) -- (4.74,0) -- cycle    ;
\draw   (455,2032.06) -- (677.5,2032.06) -- (677.5,2176.06) -- (455,2176.06) -- cycle ;
\draw [color={rgb, 255:red, 128; green, 128; blue, 128 }  ,draw opacity=1 ][fill={rgb, 255:red, 155; green, 155; blue, 155 }  ,fill opacity=1 ]   (566.67,2007.27) -- (566.67,2029.06) ;
\draw [shift={(566.67,2032.06)}, rotate = 270] [fill={rgb, 255:red, 128; green, 128; blue, 128 }  ,fill opacity=1 ][line width=0.08]  [draw opacity=0] (7.14,-3.43) -- (0,0) -- (7.14,3.43) -- (4.74,0) -- cycle    ;
\draw  [color={rgb, 255:red, 128; green, 128; blue, 128 }  ,draw opacity=1 ][dash pattern={on 4.5pt off 4.5pt}] (408.77,1691.63) .. controls (408.77,1682.47) and (416.2,1675.04) .. (425.36,1675.04) -- (706.41,1675.04) .. controls (715.57,1675.04) and (723,1682.47) .. (723,1691.63) -- (723,2177.97) .. controls (723,2187.13) and (715.57,2194.56) .. (706.41,2194.56) -- (425.36,2194.56) .. controls (416.2,2194.56) and (408.77,2187.13) .. (408.77,2177.97) -- cycle ;
\draw   (187.58,1699.76) -- (345.45,1699.76) -- (324.91,1756.31) -- (167.05,1756.31) -- cycle ;
\draw [color={rgb, 255:red, 128; green, 128; blue, 128 }  ,draw opacity=1 ][fill={rgb, 255:red, 155; green, 155; blue, 155 }  ,fill opacity=1 ]   (336,2146.31) -- (409,2146.31) ;
\draw [shift={(333,2146.31)}, rotate = 0] [fill={rgb, 255:red, 128; green, 128; blue, 128 }  ,fill opacity=1 ][line width=0.08]  [draw opacity=0] (7.14,-3.43) -- (0,0) -- (7.14,3.43) -- (4.74,0) -- cycle    ;
\draw   (186.21,2117.8) -- (344.1,2117.8) -- (323.56,2174.35) -- (165.67,2174.35) -- cycle ;

\draw (565.71,1711.23) node   [align=left] {\begin{minipage}[lt]{151.16pt}\setlength\topsep{0pt}
\begin{center}
\textbf{{\tiny Data Preprocessing}}
\end{center}

\end{minipage}};
\draw  [color={rgb, 255:red, 255; green, 255; blue, 255 }  ,draw opacity=1 ][fill={rgb, 255:red, 240; green, 240; blue, 240 }  ,fill opacity=1 ]  (462.44,1724.98) -- (669.44,1724.98) -- (669.44,1756.98) -- (462.44,1756.98) -- cycle  ;
\draw (565.94,1740.98) node  [font=\tiny] [align=left] {\begin{minipage}[lt]{138.46pt}\setlength\topsep{0pt}
\begin{center}
Sentence tokenization
\end{center}

\end{minipage}};
\draw  [color={rgb, 255:red, 255; green, 255; blue, 255 }  ,draw opacity=1 ][fill={rgb, 255:red, 240; green, 240; blue, 240 }  ,fill opacity=1 ]  (462.46,1761.98) -- (669.46,1761.98) -- (669.46,1793.98) -- (462.46,1793.98) -- cycle  ;
\draw (565.96,1777.98) node  [font=\tiny] [align=left] {\begin{minipage}[lt]{138.46pt}\setlength\topsep{0pt}
\begin{center}
Unique sentence acquisition
\end{center}

\end{minipage}};
\draw  [color={rgb, 255:red, 255; green, 255; blue, 255 }  ,draw opacity=1 ][fill={rgb, 255:red, 240; green, 240; blue, 240 }  ,fill opacity=1 ]  (462.46,1797.98) -- (669.46,1797.98) -- (669.46,1829.98) -- (462.46,1829.98) -- cycle  ;
\draw (565.96,1813.98) node  [font=\tiny] [align=left] {\begin{minipage}[lt]{138.46pt}\setlength\topsep{0pt}
\begin{center}
Short sentence removal
\end{center}

\end{minipage}};
\draw  [color={rgb, 255:red, 255; green, 255; blue, 255 }  ,draw opacity=1 ][fill={rgb, 255:red, 240; green, 240; blue, 240 }  ,fill opacity=1 ]  (431.69,1893.98) -- (700.69,1893.98) -- (700.69,1925.98) -- (431.69,1925.98) -- cycle  ;
\draw (566.19,1909.98) node  [font=\tiny] [align=left] {\begin{minipage}[lt]{180.46pt}\setlength\topsep{0pt}
\begin{center}
Sentence Embedding (SimCSE)
\end{center}

\end{minipage}};
\draw (566.02,1880.56) node  [font=\tiny] [align=left] {\begin{minipage}[lt]{192.24pt}\setlength\topsep{0pt}
\begin{center}
\tiny \textbf{Semantic Clustering}
\end{center}

\end{minipage}};
\draw  [color={rgb, 255:red, 255; green, 255; blue, 255 }  ,draw opacity=1 ][fill={rgb, 255:red, 240; green, 240; blue, 240 }  ,fill opacity=1 ]  (431.89,1930.98) -- (700.89,1930.98) -- (700.89,1962.98) -- (431.89,1962.98) -- cycle  ;
\draw (566.39,1946.98) node  [font=\tiny] [align=left] {\begin{minipage}[lt]{180.46pt}\setlength\topsep{0pt}
\begin{center}
Dimension Reduction (UMAP)
\end{center}

\end{minipage}};
\draw  [color={rgb, 255:red, 255; green, 255; blue, 255 }  ,draw opacity=1 ][fill={rgb, 255:red, 240; green, 240; blue, 240 }  ,fill opacity=1 ]  (431.91,1967.98) -- (700.91,1967.98) -- (700.91,1999.98) -- (431.91,1999.98) -- cycle  ;
\draw (566.41,1983.98) node  [font=\tiny] [align=left] {\begin{minipage}[lt]{180.46pt}\setlength\topsep{0pt}
\begin{center}
Sentence-density clustering (HDBSCAN)
\end{center}

\end{minipage}};
\draw (566.21,2050.23) node  [font=\tiny] [align=left] {\begin{minipage}[lt]{151.3pt}\setlength\topsep{0pt}
\begin{center}
{\tiny \textbf{Automatic Cluster Labelling}}
\end{center}

\end{minipage}};
\draw  [color={rgb, 255:red, 255; green, 255; blue, 255 }  ,draw opacity=1 ][fill={rgb, 255:red, 240; green, 240; blue, 240 }  ,fill opacity=1 ]  (461.53,2063.98) -- (670.53,2063.98) -- (670.53,2095.98) -- (461.53,2095.98) -- cycle  ;
\draw (566.03,2079.98) node  [font=\tiny] [align=left] {\begin{minipage}[lt]{139.36pt}\setlength\topsep{0pt}
\begin{center}
Word segmentation
\end{center}

\end{minipage}};
\draw  [color={rgb, 255:red, 255; green, 255; blue, 255 }  ,draw opacity=1 ][fill={rgb, 255:red, 240; green, 240; blue, 240 }  ,fill opacity=1 ]  (461.95,2100.48) -- (670.95,2100.48) -- (670.95,2132.48) -- (461.95,2132.48) -- cycle  ;
\draw (566.45,2116.48) node  [font=\tiny] [align=left] {\begin{minipage}[lt]{139.36pt}\setlength\topsep{0pt}
\begin{center}
POS and Dependency Tagging\\
\end{center}

\end{minipage}};
\draw  [color={rgb, 255:red, 255; green, 255; blue, 255 }  ,draw opacity=1 ][fill={rgb, 255:red, 240; green, 240; blue, 240 }  ,fill opacity=1 ]  (461.84,2136.98) -- (670.84,2136.98) -- (670.84,2168.98) -- (461.84,2168.98) -- cycle  ;
\draw (566.34,2152.98) node  [font=\tiny] [align=left] {\begin{minipage}[lt]{139.36pt}\setlength\topsep{0pt}
\begin{center}
Rule-based tag extraction
\end{center}

\end{minipage}};
\draw (565.77,2215.31) node  [font=\tiny] [align=left] {\begin{minipage}[lt]{213.66pt}\setlength\topsep{0pt}
\begin{center}
\textit{Open Entity Discovery Framework}
\end{center}

\end{minipage}};
\draw (257.22,1728.04) node  [font=\tiny] [align=left] {\begin{minipage}[lt]{119.99pt}\setlength\topsep{0pt}
\begin{center}
\textbf{{\tiny FAQ Data}}
\end{center}

\end{minipage}};
\draw (375.5,1715.52) node  [font=\tiny] [align=left] {\begin{minipage}[lt]{36.04pt}\setlength\topsep{0pt}
\begin{center}
\textit{{\tiny Input}}
\end{center}

\end{minipage}};
\draw (254.55,2146.08) node  [font=\tiny] [align=left] {\begin{minipage}[lt]{121.79pt}\setlength\topsep{0pt}
\begin{center}
{\tiny \textbf{Extracted Intentions}}
\end{center}

\end{minipage}};
\draw (375.5,2133.56) node  [font=\tiny] [align=left] {\begin{minipage}[lt]{36.04pt}\setlength\topsep{0pt}
\begin{center}
{\tiny \textit{Output}}
\end{center}

\end{minipage}};

\end{tikzpicture}

%% file: latex_figures/figure_07_list_intentions.tex
\tikzset{every picture/.style={line width=0.75pt}} 
\hspace*{-1.4cm}\begin{tikzpicture}[scale=0.179,x=1.75pt,y=1.75pt,yscale=-1,xscale=1]

\draw [color={rgb, 255:red, 128; green, 128; blue, 128 }  ,draw opacity=1 ]   (785,2108) -- (784.67,2128.76) ;
\draw [color={rgb, 255:red, 128; green, 128; blue, 128 }  ,draw opacity=1 ]   (756,2096.5) -- (785.5,2107.91) ;
\draw  [color={rgb, 255:red, 0; green, 0; blue, 0 }  ,draw opacity=1 ][fill={rgb, 255:red, 240; green, 240; blue, 240 }  ,fill opacity=1 ][line width=0.75]  (778,2107.91) .. controls (778,2103.77) and (781.36,2100.41) .. (785.5,2100.41) .. controls (789.65,2100.41) and (793,2103.77) .. (793,2107.91) .. controls (793,2112.06) and (789.65,2115.41) .. (785.5,2115.41) .. controls (781.36,2115.41) and (778,2112.06) .. (778,2107.91) -- cycle ;
\draw [color={rgb, 255:red, 128; green, 128; blue, 128 }  ,draw opacity=1 ]   (523.67,2132.76) -- (482,2163) ;
\draw [color={rgb, 255:red, 128; green, 128; blue, 128 }  ,draw opacity=1 ]   (522.83,2132.92) -- (539.5,2149.59) ;
\draw [color={rgb, 255:red, 128; green, 128; blue, 128 }  ,draw opacity=1 ]   (146.5,2262) -- (171,2261.84) ;
\draw [color={rgb, 255:red, 128; green, 128; blue, 128 }  ,draw opacity=1 ]   (178.99,2149.5) -- (145.49,2171.09) ;
\draw [color={rgb, 255:red, 128; green, 128; blue, 128 }  ,draw opacity=1 ]   (202.24,2162.09) -- (178.99,2149.5) ;
\draw [color={rgb, 255:red, 128; green, 128; blue, 128 }  ,draw opacity=1 ]   (201.49,2135.84) -- (178.99,2149.5) ;
\draw [color={rgb, 255:red, 128; green, 128; blue, 128 }  ,draw opacity=1 ]   (163.49,2126.09) -- (178.99,2149.5) ;
\draw  [color={rgb, 255:red, 0; green, 0; blue, 0 }  ,draw opacity=1 ][fill={rgb, 255:red, 240; green, 240; blue, 240 }  ,fill opacity=1 ][line width=0.75]  (171.49,2149.5) .. controls (171.49,2145.36) and (174.85,2142) .. (178.99,2142) .. controls (183.13,2142) and (186.49,2145.36) .. (186.49,2149.5) .. controls (186.49,2153.64) and (183.13,2157) .. (178.99,2157) .. controls (174.85,2157) and (171.49,2153.64) .. (171.49,2149.5) -- cycle ;
\draw [color={rgb, 255:red, 128; green, 128; blue, 128 }  ,draw opacity=1 ]   (357.5,2120) -- (357.5,2141.34) ;
\draw [color={rgb, 255:red, 128; green, 128; blue, 128 }  ,draw opacity=1 ]   (356.67,2098.76) -- (357.5,2120) ;
\draw  [color={rgb, 255:red, 0; green, 0; blue, 0 }  ,draw opacity=1 ][fill={rgb, 255:red, 240; green, 240; blue, 240 }  ,fill opacity=1 ][line width=0.75]  (350,2120) .. controls (350,2115.86) and (353.36,2112.5) .. (357.5,2112.5) .. controls (361.65,2112.5) and (365,2115.86) .. (365,2120) .. controls (365,2124.14) and (361.65,2127.5) .. (357.5,2127.5) .. controls (353.36,2127.5) and (350,2124.14) .. (350,2120) -- cycle ;
\draw [color={rgb, 255:red, 128; green, 128; blue, 128 }  ,draw opacity=1 ]   (146.5,2263) -- (157,2284) ;
\draw [color={rgb, 255:red, 128; green, 128; blue, 128 }  ,draw opacity=1 ]   (147,2242.59) -- (146.5,2262) ;
\draw  [color={rgb, 255:red, 0; green, 0; blue, 0 }  ,draw opacity=1 ][fill={rgb, 255:red, 240; green, 240; blue, 240 }  ,fill opacity=1 ][line width=0.75]  (139,2263) .. controls (139,2258.86) and (142.36,2255.5) .. (146.5,2255.5) .. controls (150.65,2255.5) and (154,2258.86) .. (154,2263) .. controls (154,2267.14) and (150.65,2270.5) .. (146.5,2270.5) .. controls (142.36,2270.5) and (139,2267.14) .. (139,2263) -- cycle ;
\draw [color={rgb, 255:red, 128; green, 128; blue, 128 }  ,draw opacity=1 ]   (556.13,2272) -- (582,2268.59) ;
\draw [color={rgb, 255:red, 128; green, 128; blue, 128 }  ,draw opacity=1 ]   (556.13,2272) -- (538,2296.09) ;
\draw [color={rgb, 255:red, 128; green, 128; blue, 128 }  ,draw opacity=1 ]   (527,2255.59) -- (556.13,2272) ;
\draw  [color={rgb, 255:red, 0; green, 0; blue, 0 }  ,draw opacity=1 ][fill={rgb, 255:red, 240; green, 240; blue, 240 }  ,fill opacity=1 ][line width=0.75]  (548.63,2272) .. controls (548.63,2267.86) and (551.99,2264.5) .. (556.13,2264.5) .. controls (560.27,2264.5) and (563.63,2267.86) .. (563.63,2272) .. controls (563.63,2276.14) and (560.27,2279.5) .. (556.13,2279.5) .. controls (551.99,2279.5) and (548.63,2276.14) .. (548.63,2272) -- cycle ;
\draw [color={rgb, 255:red, 128; green, 128; blue, 128 }  ,draw opacity=1 ]   (523.13,2131) -- (549.67,2128.09) ;
\draw [color={rgb, 255:red, 128; green, 128; blue, 128 }  ,draw opacity=1 ]   (523.13,2131) -- (492.67,2128.76) ;
\draw [color={rgb, 255:red, 128; green, 128; blue, 128 }  ,draw opacity=1 ]   (529.83,2101.26) -- (523.13,2131) ;
\draw  [color={rgb, 255:red, 0; green, 0; blue, 0 }  ,draw opacity=1 ][fill={rgb, 255:red, 240; green, 240; blue, 240 }  ,fill opacity=1 ][line width=0.75]  (515.63,2131) .. controls (515.63,2126.86) and (518.99,2123.5) .. (523.13,2123.5) .. controls (527.27,2123.5) and (530.63,2126.86) .. (530.63,2131) .. controls (530.63,2135.14) and (527.27,2138.5) .. (523.13,2138.5) .. controls (518.99,2138.5) and (515.63,2135.14) .. (515.63,2131) -- cycle ;
\draw [color={rgb, 255:red, 128; green, 128; blue, 128 }  ,draw opacity=1 ]   (371.76,2250) -- (396.13,2223.59) ;
\draw [color={rgb, 255:red, 128; green, 128; blue, 128 }  ,draw opacity=1 ]   (371.76,2250) -- (364.79,2269.92) ;
\draw [color={rgb, 255:red, 128; green, 128; blue, 128 }  ,draw opacity=1 ]   (342.13,2217.26) -- (371.76,2250) ;
\draw  [color={rgb, 255:red, 0; green, 0; blue, 0 }  ,draw opacity=1 ][fill={rgb, 255:red, 240; green, 240; blue, 240 }  ,fill opacity=1 ][line width=0.75]  (364.26,2250) .. controls (364.26,2245.86) and (367.62,2242.5) .. (371.76,2242.5) .. controls (375.9,2242.5) and (379.26,2245.86) .. (379.26,2250) .. controls (379.26,2254.14) and (375.9,2257.5) .. (371.76,2257.5) .. controls (367.62,2257.5) and (364.26,2254.14) .. (364.26,2250) -- cycle ;
\draw [color={rgb, 255:red, 128; green, 128; blue, 128 }  ,draw opacity=1 ]   (725.5,2223.09) -- (756.48,2231.59) ;
\draw [color={rgb, 255:red, 128; green, 128; blue, 128 }  ,draw opacity=1 ]   (726.5,2225.59) -- (746.48,2201.59) ;
\draw [color={rgb, 255:red, 128; green, 128; blue, 128 }  ,draw opacity=1 ]   (727.05,2225.5) -- (736,2257.5) ;
\draw [color={rgb, 255:red, 128; green, 128; blue, 128 }  ,draw opacity=1 ]   (701.5,2223.59) -- (727.05,2225.5) ;
\draw  [color={rgb, 255:red, 0; green, 0; blue, 0 }  ,draw opacity=1 ][fill={rgb, 255:red, 240; green, 240; blue, 240 }  ,fill opacity=1 ][line width=0.75]  (719.55,2225.5) .. controls (719.55,2221.36) and (722.91,2218) .. (727.05,2218) .. controls (731.19,2218) and (734.55,2221.36) .. (734.55,2225.5) .. controls (734.55,2229.64) and (731.19,2233) .. (727.05,2233) .. controls (722.91,2233) and (719.55,2229.64) .. (719.55,2225.5) -- cycle ;

\draw (780.5,2136) node   [align=left] {\begin{minipage}[lt]{134.76pt}\setlength\topsep{0pt}
\begin{center} \tiny
make-up exam arrangement
\end{center}

\end{minipage}};
\draw (686.5,2094) node   [align=left] {\begin{minipage}[lt]{98.04pt}\setlength\topsep{0pt}
\begin{center} \tiny
exam postponement
\end{center}

\end{minipage}};
\draw (164.49,2113.5) node   [align=left] {\begin{minipage}[lt]{88.41pt}\setlength\topsep{0pt}
\begin{center} \tiny
grade adjustment
\end{center}

\end{minipage}};
\draw (247,2132.61) node   [align=left] {\begin{minipage}[lt]{88.41pt}\setlength\topsep{0pt}
\begin{center} \tiny
grade review
\end{center}

\end{minipage}};
\draw (249,2161.89) node   [align=left] {\begin{minipage}[lt]{88.41pt}\setlength\topsep{0pt}
\begin{center} \tiny
grade inquiry
\end{center}

\end{minipage}};
\draw (146.49,2182.5) node   [align=left] {\begin{minipage}[lt]{74.81pt}\setlength\topsep{0pt}
\begin{center} \tiny
grade checking
\end{center}

\end{minipage}};
\draw (357,2090.5) node   [align=left] {\begin{minipage}[lt]{88.41pt}\setlength\topsep{0pt}
\begin{center} \tiny
study consultation
\end{center}

\end{minipage}};
\draw (358,2148.5) node   [align=left] {\begin{minipage}[lt]{74.81pt}\setlength\topsep{0pt}
\begin{center} \tiny
study materials
\end{center}

\end{minipage}};
\draw (146.5,2231.5) node   [align=left] {\begin{minipage}[lt]{102.35pt}\setlength\topsep{0pt}
\begin{center} \tiny
class change support
\end{center}

\end{minipage}};
\draw (158.5,2295.5) node   [align=left] {\begin{minipage}[lt]{102.85pt}\setlength\topsep{0pt}
\begin{center} \tiny
class addition support
\end{center}

\end{minipage}};
\draw (244.5,2260.5) node   [align=left] {\begin{minipage}[lt]{102.85pt}\setlength\topsep{0pt}
\begin{center} \tiny
class transfer support
\end{center}

\end{minipage}};
\draw (525.5,2245.5) node   [align=left] {\begin{minipage}[lt]{87.81pt}\setlength\topsep{0pt}
\begin{center} \tiny
course withdrawal
\end{center}

\end{minipage}};
\draw (624,2266.5) node   [align=left] {\begin{minipage}[lt]{61.24pt}\setlength\topsep{0pt}
\begin{center} \tiny
course drop
\end{center}

\end{minipage}};
\draw (536,2304.5) node   [align=left] {\begin{minipage}[lt]{144.2pt}\setlength\topsep{0pt}
\begin{center} \tiny
course withdrawal registration
\end{center}

\end{minipage}};
\draw (530,2092.5) node   [align=left] {\begin{minipage}[lt]{86.74pt}\setlength\topsep{0pt}
\begin{center} \tiny
subject schedule
\end{center}

\end{minipage}};
\draw (604.5,2127) node   [align=left] {\begin{minipage}[lt]{81.28pt}\setlength\topsep{0pt}
\begin{center} \tiny
timetable check
\end{center}

\end{minipage}};
\draw (611,2154.5) node   [align=left] {\begin{minipage}[lt]{103.04pt}\setlength\topsep{0pt}
\begin{center} \tiny
study schedule check
\end{center}

\end{minipage}};
\draw (456,2127.5) node   [align=left] {\begin{minipage}[lt]{54.58pt}\setlength\topsep{0pt}
\begin{center} \tiny
study plan
\end{center}

\end{minipage}};
\draw (482,2173.5) node   [align=left] {\begin{minipage}[lt]{103.04pt}\setlength\topsep{0pt}
\begin{center} \tiny
overlapping schedule
\end{center}

\end{minipage}};
\draw (342,2205.5) node   [align=left] {\begin{minipage}[lt]{66.81pt}\setlength\topsep{0pt}
\begin{center} \tiny
class opening
\end{center}

\end{minipage}};
\draw (365.63,2277) node   [align=left] {\begin{minipage}[lt]{88.83pt}\setlength\topsep{0pt}
\begin{center} \tiny
new class opening
\end{center}

\end{minipage}};
\draw (457.63,2219) node   [align=left] {\begin{minipage}[lt]{95.63pt}\setlength\topsep{0pt}
\begin{center} \tiny
class size increase
\end{center}

\end{minipage}};
\draw (736,2267.5) node   [align=left] {\begin{minipage}[lt]{84.38pt}\setlength\topsep{0pt}
\begin{center} \tiny
class registration
\end{center}

\end{minipage}};
\draw (636.5,2222) node   [align=left] {\begin{minipage}[lt]{88.43pt}\setlength\topsep{0pt}
\begin{center} \tiny
course registration
\end{center}

\end{minipage}};
\draw (746.48,2192) node   [align=left] {\begin{minipage}[lt]{131.98pt}\setlength\topsep{0pt}
\begin{center} \tiny
course registration deadline
\end{center}

\end{minipage}};
\draw (822.48,2230) node   [align=left] {\begin{minipage}[lt]{91.15pt}\setlength\topsep{0pt}
\begin{center} \tiny
subject registration
\end{center}

\end{minipage}};

\end{tikzpicture}

%% file: latex_figures/figure_06_embedding_based_method.tex
\tikzset{every picture/.style={line width=0.75pt}} 
\hspace*{-0.9cm}\begin{tikzpicture}[scale=0.36,x=0.95pt,y=0.8pt,yscale=-1,xscale=1]

\draw [color={rgb, 255:red, 128; green, 128; blue, 128 }  ,draw opacity=1 ][fill={rgb, 255:red, 155; green, 155; blue, 155 }  ,fill opacity=1 ][line width=0.75]    (344.67,2968.8) -- (415.67,2968.8) ;
\draw [shift={(418.67,2968.8)}, rotate = 180] [fill={rgb, 255:red, 128; green, 128; blue, 128 }  ,fill opacity=1 ][line width=0.08]  [draw opacity=0] (7.14,-3.43) -- (0,0) -- (7.14,3.43) -- (4.74,0) -- cycle    ;
\draw [color={rgb, 255:red, 128; green, 128; blue, 128 }  ,draw opacity=1 ][fill={rgb, 255:red, 155; green, 155; blue, 155 }  ,fill opacity=1 ][line width=0.75]    (509.17,2928.13) -- (509.17,2949.92) ;
\draw [shift={(509.17,2952.92)}, rotate = 270] [fill={rgb, 255:red, 128; green, 128; blue, 128 }  ,fill opacity=1 ][line width=0.08]  [draw opacity=0] (7.14,-3.43) -- (0,0) -- (7.14,3.43) -- (4.74,0) -- cycle    ;
\draw [color={rgb, 255:red, 128; green, 128; blue, 128 }  ,draw opacity=1 ][fill={rgb, 255:red, 155; green, 155; blue, 155 }  ,fill opacity=1 ][line width=0.75]    (509.42,2985.46) -- (509.42,3007.25) ;
\draw [shift={(509.42,3010.25)}, rotate = 270] [fill={rgb, 255:red, 128; green, 128; blue, 128 }  ,fill opacity=1 ][line width=0.08]  [draw opacity=0] (7.14,-3.43) -- (0,0) -- (7.14,3.43) -- (4.74,0) -- cycle    ;
\draw  [color={rgb, 255:red, 128; green, 128; blue, 128 }  ,draw opacity=1 ][dash pattern={on 4.5pt off 4.5pt}][line width=0.75]  (418.43,2888.96) .. controls (418.43,2883.77) and (422.64,2879.56) .. (427.83,2879.56) -- (590.6,2879.56) .. controls (595.79,2879.56) and (600,2883.77) .. (600,2888.96) -- (600,3048.16) .. controls (600,3053.35) and (595.79,3057.56) .. (590.6,3057.56) -- (427.83,3057.56) .. controls (422.64,3057.56) and (418.43,3053.35) .. (418.43,3048.16) -- cycle ;
\draw  [line width=0.75]  (197.25,2940.29) -- (355.12,2940.29) -- (334.58,2996.84) -- (176.71,2996.84) -- cycle ;
\draw   (684.21,2940.8) -- (842.1,2940.8) -- (821.56,2997.35) -- (663.67,2997.35) -- cycle ;
\draw [color={rgb, 255:red, 128; green, 128; blue, 128 }  ,draw opacity=1 ][fill={rgb, 255:red, 155; green, 155; blue, 155 }  ,fill opacity=1 ][line width=0.75]    (599.67,2968.8) -- (670.67,2968.8) ;
\draw [shift={(673.67,2968.8)}, rotate = 180] [fill={rgb, 255:red, 128; green, 128; blue, 128 }  ,fill opacity=1 ][line width=0.08]  [draw opacity=0] (7.14,-3.43) -- (0,0) -- (7.14,3.43) -- (4.74,0) -- cycle    ;

\draw  [line width=0.75]   (430.19,2895.75) -- (589.19,2895.75) -- (589.19,2927.75) -- (430.19,2927.75) -- cycle  ;
\draw (509.69,2911.75) node  [font=\tiny] [align=left] {\begin{minipage}[lt]{105.08pt}\setlength\topsep{0pt}
\begin{center}
Entity Enrich
\end{center}

\end{minipage}};
\draw  [line width=0.75]   (430.22,2953.09) -- (589.22,2953.09) -- (589.22,2985.09) -- (430.22,2985.09) -- cycle  ;
\draw (509.72,2969.09) node  [font=\tiny] [align=left] {\begin{minipage}[lt]{105.08pt}\setlength\topsep{0pt}
\begin{center}
Entity Embedding
\end{center}

\end{minipage}};
\draw  [line width=0.75]   (430.22,3009.75) -- (589.22,3009.75) -- (589.22,3041.75) -- (430.22,3041.75) -- cycle  ;
\draw (509.72,3025.75) node  [font=\tiny] [align=left] {\begin{minipage}[lt]{105.08pt}\setlength\topsep{0pt}
\begin{center}
Similarity Measuring
\end{center}

\end{minipage}};
\draw (508.92,3077.31) node  [font=\tiny] [align=left] {\begin{minipage}[lt]{132.03pt}\setlength\topsep{0pt}
\begin{center}
\textit{Relation Discovery Module}
\end{center}

\end{minipage}};
\draw (265.91,2968.56) node  [font=\tiny] [align=left] {\begin{minipage}[lt]{119.99pt}\setlength\topsep{0pt}
\begin{center}
Intent\\Policy
\end{center}

\end{minipage}};
\draw (382.17,2956.05) node  [font=\tiny] [align=left] {\begin{minipage}[lt]{36.04pt}\setlength\topsep{0pt}
\begin{center}
\textit{{\tiny Input}}
\end{center}

\end{minipage}};
\draw (275.58,2925.54) node  [font=\tiny] [align=left] {\begin{minipage}[lt]{106.53pt}\setlength\topsep{0pt}
\begin{center}
{\tiny \textit{Entity}}
\end{center}

\end{minipage}};
\draw (752.55,2969.08) node  [font=\tiny] [align=left] {\begin{minipage}[lt]{121.79pt}\setlength\topsep{0pt}
\begin{center}
Entity Relationships
\end{center}

\end{minipage}};
\draw (637.17,2956.05) node  [font=\tiny] [align=left] {\begin{minipage}[lt]{36.04pt}\setlength\topsep{0pt}
\begin{center}
\textit{{\tiny Output}}
\end{center}

\end{minipage}};

\end{tikzpicture}

%% file: latex_figures/figure_04_kgrag_pipeline.tex
\tikzset{every picture/.style={line width=0.75pt}} 
\hspace*{-1cm}\begin{tikzpicture}[scale=0.45, x=0.73pt,y=0.73pt,yscale=-1,xscale=1]

\draw   (165.79,964.76) -- (282.36,964.76) -- (268.57,1021.31) -- (152,1021.31) -- cycle ;
\draw  [fill={rgb, 255:red, 240; green, 240; blue, 240 }  ,fill opacity=1 ] (309.64,1005.14) -- (309.64,998.58) -- (286.49,998.58) -- (286.49,985.46) -- (309.64,985.46) -- (309.64,978.9) -- (325.07,992.02) -- cycle ;
\draw   (323.21,1059.23) -- (477.18,1059.23) -- (458.97,1115.77) -- (305.01,1115.77) -- cycle ;
\draw  [fill={rgb, 255:red, 240; green, 240; blue, 240 }  ,fill opacity=1 ] (377.8,1037.82) -- (384.92,1037.7) -- (384.52,1016.39) -- (398.77,1016.16) -- (399.17,1037.48) -- (406.3,1037.36) -- (392.32,1051.8) -- cycle ;
\draw   (529.6,1063.94) -- (688.43,1063.94) -- (688.43,1111.06) -- (529.6,1111.06) -- cycle(681.36,1071.01) -- (536.67,1071.01) -- (536.67,1103.99) -- (681.36,1103.99) -- cycle ;
\draw  [fill={rgb, 255:red, 240; green, 240; blue, 240 }  ,fill opacity=1 ] (504.9,1100.55) -- (504.9,1093.99) -- (481.75,1093.99) -- (481.75,1080.87) -- (504.9,1080.87) -- (504.9,1074.31) -- (520.33,1087.43) -- cycle ;
\draw   (754.58,1059.23) -- (895.91,1059.23) -- (879.19,1115.77) -- (737.85,1115.77) -- cycle ;
\draw  [fill={rgb, 255:red, 240; green, 240; blue, 240 }  ,fill opacity=1 ] (720.95,1100.69) -- (720.95,1094.13) -- (697.8,1094.13) -- (697.8,1081.01) -- (720.95,1081.01) -- (720.95,1074.45) -- (736.38,1087.57) -- cycle ;
\draw [color={rgb, 255:red, 128; green, 128; blue, 128 }  ,draw opacity=1 ][fill={rgb, 255:red, 128; green, 128; blue, 128 }  ,fill opacity=1 ][line width=0.75]    (543.04,1022.3) -- (577.76,1014.21) ;
\draw [color={rgb, 255:red, 128; green, 128; blue, 128 }  ,draw opacity=1 ][fill={rgb, 255:red, 128; green, 128; blue, 128 }  ,fill opacity=1 ][line width=0.75]    (517.73,1015.66) -- (543.04,1022.3) ;
\draw [color={rgb, 255:red, 128; green, 128; blue, 128 }  ,draw opacity=1 ][fill={rgb, 255:red, 128; green, 128; blue, 128 }  ,fill opacity=1 ][line width=0.75]    (577.82,1018.44) -- (584.5,983.78) ;
\draw [color={rgb, 255:red, 128; green, 128; blue, 128 }  ,draw opacity=1 ][fill={rgb, 255:red, 128; green, 128; blue, 128 }  ,fill opacity=1 ][line width=0.75]    (584.5,986.55) -- (557.79,969.92) ;
\draw [color={rgb, 255:red, 128; green, 128; blue, 128 }  ,draw opacity=1 ][fill={rgb, 255:red, 128; green, 128; blue, 128 }  ,fill opacity=1 ][line width=0.75]    (570.48,995.56) -- (584.5,984.47) ;
\draw [color={rgb, 255:red, 128; green, 128; blue, 128 }  ,draw opacity=1 ][fill={rgb, 255:red, 128; green, 128; blue, 128 }  ,fill opacity=1 ][line width=0.75]    (551.12,1004.57) -- (571.14,992.79) ;
\draw [color={rgb, 255:red, 128; green, 128; blue, 128 }  ,draw opacity=1 ][fill={rgb, 255:red, 128; green, 128; blue, 128 }  ,fill opacity=1 ][line width=0.75]    (544.44,997.64) -- (564.47,969.92) ;
\draw [color={rgb, 255:red, 128; green, 128; blue, 128 }  ,draw opacity=1 ][fill={rgb, 255:red, 128; green, 128; blue, 128 }  ,fill opacity=1 ][line width=0.75]    (551.12,1007.34) -- (577.82,1014.28) ;
\draw [color={rgb, 255:red, 128; green, 128; blue, 128 }  ,draw opacity=1 ][fill={rgb, 255:red, 128; green, 128; blue, 128 }  ,fill opacity=1 ][line width=0.75]    (517.73,1015.66) -- (544.44,1004.57) ;
\draw [color={rgb, 255:red, 128; green, 128; blue, 128 }  ,draw opacity=1 ][fill={rgb, 255:red, 128; green, 128; blue, 128 }  ,fill opacity=1 ][line width=0.75]    (510.39,978.23) -- (517.73,1010.58) ;
\draw [color={rgb, 255:red, 128; green, 128; blue, 128 }  ,draw opacity=1 ][fill={rgb, 255:red, 128; green, 128; blue, 128 }  ,fill opacity=1 ][line width=0.75]    (517.73,983.78) -- (537.76,997.64) ;
\draw [color={rgb, 255:red, 128; green, 128; blue, 128 }  ,draw opacity=1 ][fill={rgb, 255:red, 128; green, 128; blue, 128 }  ,fill opacity=1 ][line width=0.75]    (532.42,967.61) -- (511.06,976.85) ;
\draw  [fill={rgb, 255:red, 240; green, 240; blue, 240 }  ,fill opacity=1 ] (472.43,978.74) -- (472.43,985.3) -- (495.58,985.3) -- (495.58,998.43) -- (472.43,998.43) -- (472.43,1004.99) -- (457,991.86) -- cycle ;
\draw  [color={rgb, 255:red, 128; green, 128; blue, 128 }  ,draw opacity=1 ][fill={rgb, 255:red, 255; green, 255; blue, 255 }  ,fill opacity=1 ][line width=0.75]  (504.38,981.47) .. controls (504.38,976.36) and (508.57,972.23) .. (513.73,972.23) .. controls (518.89,972.23) and (523.07,976.36) .. (523.07,981.47) .. controls (523.07,986.57) and (518.89,990.71) .. (513.73,990.71) .. controls (508.57,990.71) and (504.38,986.57) .. (504.38,981.47) -- cycle ;
\draw  [color={rgb, 255:red, 128; green, 128; blue, 128 }  ,draw opacity=1 ][fill={rgb, 255:red, 255; green, 255; blue, 255 }  ,fill opacity=1 ][line width=0.75]  (527.75,999.72) .. controls (527.75,992.06) and (534.02,985.86) .. (541.77,985.86) .. controls (549.51,985.86) and (555.79,992.06) .. (555.79,999.72) .. controls (555.79,1007.38) and (549.51,1013.58) .. (541.77,1013.58) .. controls (534.02,1013.58) and (527.75,1007.38) .. (527.75,999.72) -- cycle ;
\draw  [color={rgb, 255:red, 128; green, 128; blue, 128 }  ,draw opacity=1 ][fill={rgb, 255:red, 255; green, 255; blue, 255 }  ,fill opacity=1 ][line width=0.75]  (513.73,1013.81) .. controls (513.73,1011.26) and (515.82,1009.19) .. (518.4,1009.19) .. controls (520.98,1009.19) and (523.07,1011.26) .. (523.07,1013.81) .. controls (523.07,1016.37) and (520.98,1018.44) .. (518.4,1018.44) .. controls (515.82,1018.44) and (513.73,1016.37) .. (513.73,1013.81) -- cycle ;
\draw  [color={rgb, 255:red, 128; green, 128; blue, 128 }  ,draw opacity=1 ][fill={rgb, 255:red, 255; green, 255; blue, 255 }  ,fill opacity=1 ][line width=0.75]  (579.16,986.09) .. controls (579.16,983.54) and (581.25,981.47) .. (583.83,981.47) .. controls (586.41,981.47) and (588.5,983.54) .. (588.5,986.09) .. controls (588.5,988.64) and (586.41,990.71) .. (583.83,990.71) .. controls (581.25,990.71) and (579.16,988.64) .. (579.16,986.09) -- cycle ;
\draw  [color={rgb, 255:red, 128; green, 128; blue, 128 }  ,draw opacity=1 ][fill={rgb, 255:red, 255; green, 255; blue, 255 }  ,fill opacity=1 ][line width=0.75]  (551.12,972.23) .. controls (551.12,967.12) and (555.3,962.98) .. (560.46,962.98) .. controls (565.62,962.98) and (569.81,967.12) .. (569.81,972.23) .. controls (569.81,977.33) and (565.62,981.47) .. (560.46,981.47) .. controls (555.3,981.47) and (551.12,977.33) .. (551.12,972.23) -- cycle ;
\draw  [color={rgb, 255:red, 128; green, 128; blue, 128 }  ,draw opacity=1 ][fill={rgb, 255:red, 255; green, 255; blue, 255 }  ,fill opacity=1 ][line width=0.75]  (532.42,967.61) .. controls (532.42,965.05) and (534.51,962.98) .. (537.09,962.98) .. controls (539.68,962.98) and (541.77,965.05) .. (541.77,967.61) .. controls (541.77,970.16) and (539.68,972.23) .. (537.09,972.23) .. controls (534.51,972.23) and (532.42,970.16) .. (532.42,967.61) -- cycle ;
\draw [color={rgb, 255:red, 128; green, 128; blue, 128 }  ,draw opacity=1 ][fill={rgb, 255:red, 128; green, 128; blue, 128 }  ,fill opacity=1 ][line width=0.75]    (541.77,967.61) -- (551.12,969.92) ;
\draw  [color={rgb, 255:red, 128; green, 128; blue, 128 }  ,draw opacity=1 ][fill={rgb, 255:red, 255; green, 255; blue, 255 }  ,fill opacity=1 ][line width=0.75]  (571.14,1013.81) .. controls (571.14,1008.71) and (575.33,1004.57) .. (580.49,1004.57) .. controls (585.65,1004.57) and (589.84,1008.71) .. (589.84,1013.81) .. controls (589.84,1018.92) and (585.65,1023.06) .. (580.49,1023.06) .. controls (575.33,1023.06) and (571.14,1018.92) .. (571.14,1013.81) -- cycle ;
\draw  [color={rgb, 255:red, 128; green, 128; blue, 128 }  ,draw opacity=1 ][fill={rgb, 255:red, 255; green, 255; blue, 255 }  ,fill opacity=1 ][line width=0.75]  (564.47,995.33) .. controls (564.47,992.78) and (566.56,990.71) .. (569.14,990.71) .. controls (571.72,990.71) and (573.81,992.78) .. (573.81,995.33) .. controls (573.81,997.88) and (571.72,999.95) .. (569.14,999.95) .. controls (566.56,999.95) and (564.47,997.88) .. (564.47,995.33) -- cycle ;
\draw  [color={rgb, 255:red, 128; green, 128; blue, 128 }  ,draw opacity=1 ][fill={rgb, 255:red, 255; green, 255; blue, 255 }  ,fill opacity=1 ][line width=0.75]  (538.37,1022.3) .. controls (538.37,1019.74) and (540.46,1017.67) .. (543.04,1017.67) .. controls (545.62,1017.67) and (547.72,1019.74) .. (547.72,1022.3) .. controls (547.72,1024.85) and (545.62,1026.92) .. (543.04,1026.92) .. controls (540.46,1026.92) and (538.37,1024.85) .. (538.37,1022.3) -- cycle ;

\draw  [fill={rgb, 255:red, 255; green, 255; blue, 255 }  ,fill opacity=1 ]  (331.91,976.07) -- (450.91,976.07) -- (450.91,1008.07) -- (331.91,1008.07) -- cycle  ;
\draw (391.41,992.07) node   [align=left] {\begin{minipage}[lt]{78.51pt}\setlength\topsep{0pt}
\begin{center} \tiny
KG - Retrieval
\end{center}

\end{minipage}};
\draw (391.54,960.91) node  [font=\small] [align=left] {\begin{minipage}[lt]{83.7pt}\setlength\topsep{0pt}
\begin{center} \tiny
\textit{Retriever}
\end{center}

\end{minipage}};
\draw (219,993.03) node   [align=left] {\begin{minipage}[lt]{86.73pt}\setlength\topsep{0pt}
\begin{center} \tiny
User Question\\Utterance
\end{center}

\end{minipage}};
\draw (392.38,1087.5) node   [align=left] {\begin{minipage}[lt]{114.78pt}\setlength\topsep{0pt}
\begin{center} \tiny
Relevant Knowledge
\end{center}

\end{minipage}};
\draw (609.02,1087.5) node   [align=left] {\begin{minipage}[lt]{108pt}\setlength\topsep{0pt}
\begin{center} \tiny
URA LLM
\end{center}

\end{minipage}};
\draw (609.54,1048.8) node  [font=\small] [align=left] {\begin{minipage}[lt]{112.14pt}\setlength\topsep{0pt}
\begin{center} \tiny
\textit{Large Language Model}
\end{center}

\end{minipage}};
\draw (816.96,1087.5) node   [align=left] {\begin{minipage}[lt]{110.1pt}\setlength\topsep{0pt}
\begin{center} \tiny
Generated Answer
\end{center}

\end{minipage}};
\draw (643.62,970.5) node   [align=left] {\begin{minipage}[lt]{95.72pt}\setlength\topsep{0pt}
\begin{center} \tiny
\hyperref[fig:figure_02_education_knowledge_graph]{\textbf{Knowledge Graph}}
\end{center}

\end{minipage}};

\end{tikzpicture}

%% file: main.bbl

\begin{thebibliography}{30}


\ifx \showCODEN    \undefined \def \showCODEN     #1{\unskip}     \fi
\ifx \showDOI      \undefined \def \showDOI       #1{#1}\fi
\ifx \showISBNx    \undefined \def \showISBNx     #1{\unskip}     \fi
\ifx \showISBNxiii \undefined \def \showISBNxiii  #1{\unskip}     \fi
\ifx \showISSN     \undefined \def \showISSN      #1{\unskip}     \fi
\ifx \showLCCN     \undefined \def \showLCCN      #1{\unskip}     \fi
\ifx \shownote     \undefined \def \shownote      #1{#1}          \fi
\ifx \showarticletitle \undefined \def \showarticletitle #1{#1}   \fi
\ifx \showURL      \undefined \def \showURL       {\relax}        \fi
\providecommand\bibfield[2]{#2}
\providecommand\bibinfo[2]{#2}
\providecommand\natexlab[1]{#1}
\providecommand\showeprint[2][]{arXiv:#2}

\bibitem[Baek et~al\mbox{.}(2023)]%
        {baek-etal-2023-knowledge}
\bibfield{author}{\bibinfo{person}{Jinheon Baek}, \bibinfo{person}{Alham~Fikri Aji}, {and} \bibinfo{person}{Amir Saffari}.} \bibinfo{year}{2023}\natexlab{}.
\newblock \showarticletitle{Knowledge-Augmented Language Model Prompting for Zero-Shot Knowledge Graph Question Answering}. In \bibinfo{booktitle}{\emph{Proceedings of the 1st Workshop on Natural Language Reasoning and Structured Explanations (NLRSE)}}, \bibfield{editor}{\bibinfo{person}{Bhavana Dalvi~Mishra}, \bibinfo{person}{Greg Durrett}, \bibinfo{person}{Peter Jansen}, \bibinfo{person}{Danilo Neves~Ribeiro}, {and} \bibinfo{person}{Jason Wei}} (Eds.). \bibinfo{publisher}{Association for Computational Linguistics}, \bibinfo{address}{Toronto, Canada}, \bibinfo{pages}{78--106}.
\newblock
\urldef\tempurl%
\url{https://doi.org/10.18653/v1/2023.nlrse-1.7}
\showDOI{\tempurl}


\bibitem[Casanueva et~al\mbox{.}(2020)]%
        {casanueva2020efficient}
\bibfield{author}{\bibinfo{person}{Iñigo Casanueva}, \bibinfo{person}{Tadas Temčinas}, \bibinfo{person}{Daniela Gerz}, \bibinfo{person}{Matthew Henderson}, {and} \bibinfo{person}{Ivan Vulić}.} \bibinfo{year}{2020}\natexlab{}.
\newblock \bibinfo{title}{Efficient Intent Detection with Dual Sentence Encoders}.
\newblock
\newblock
\showeprint[arxiv]{2003.04807}~[cs.CL]


\bibitem[Chen et~al\mbox{.}(2022)]%
        {Chen2022IntentDF}
\bibfield{author}{\bibinfo{person}{Minhua Chen}, \bibinfo{person}{Badrinath Jayakumar}, \bibinfo{person}{Michael Johnston}, \bibinfo{person}{S.~Eman Mahmoodi}, {and} \bibinfo{person}{Daniel Pressel}.} \bibinfo{year}{2022}\natexlab{}.
\newblock \showarticletitle{Intent Discovery for Enterprise Virtual Assistants: Applications of Utterance Embedding and Clustering to Intent Mining}. In \bibinfo{booktitle}{\emph{North American Chapter of the Association for Computational Linguistics}}.
\newblock
\urldef\tempurl%
\url{https://api.semanticscholar.org/CorpusID:250390998}
\showURL{%
\tempurl}


\bibitem[Chen et~al\mbox{.}(2018)]%
        {Chen_20218}
\bibfield{author}{\bibinfo{person}{Penghe Chen}, \bibinfo{person}{Yu Lu}, \bibinfo{person}{Vincent~W. Zheng}, \bibinfo{person}{Xiyang Chen}, {and} \bibinfo{person}{Boda Yang}.} \bibinfo{year}{2018}\natexlab{}.
\newblock \showarticletitle{KnowEdu: A System to Construct Knowledge Graph for Education}.
\newblock \bibinfo{journal}{\emph{IEEE Access}}  \bibinfo{volume}{6} (\bibinfo{year}{2018}), \bibinfo{pages}{31553--31563}.
\newblock
\urldef\tempurl%
\url{https://doi.org/10.1109/ACCESS.2018.2839607}
\showDOI{\tempurl}


\bibitem[Fettach et~al\mbox{.}(2022)]%
        {Fettach_2022}
\bibfield{author}{\bibinfo{person}{Yousra Fettach}, \bibinfo{person}{Mounir Ghogho}, {and} \bibinfo{person}{Boualem Benatallah}.} \bibinfo{year}{2022}\natexlab{}.
\newblock \showarticletitle{Knowledge Graphs in Education and Employability: A Survey on Applications and Techniques}.
\newblock \bibinfo{journal}{\emph{IEEE Access}}  \bibinfo{volume}{10} (\bibinfo{year}{2022}), \bibinfo{pages}{80174--80183}.
\newblock
\showISSN{2169-3536}
\urldef\tempurl%
\url{https://doi.org/10.1109/ACCESS.2022.3194063}
\showDOI{\tempurl}


\bibitem[Gao et~al\mbox{.}(2022)]%
        {gao2022simcse}
\bibfield{author}{\bibinfo{person}{Tianyu Gao}, \bibinfo{person}{Xingcheng Yao}, {and} \bibinfo{person}{Danqi Chen}.} \bibinfo{year}{2022}\natexlab{}.
\newblock \bibinfo{title}{SimCSE: Simple Contrastive Learning of Sentence Embeddings}.
\newblock
\newblock
\showeprint[arxiv]{2104.08821}~[cs.CL]


\bibitem[Grootendorst(2022)]%
        {grootendorst2022bertopic}
\bibfield{author}{\bibinfo{person}{Maarten Grootendorst}.} \bibinfo{year}{2022}\natexlab{}.
\newblock \showarticletitle{BERTopic: Neural topic modeling with a class-based TF-IDF procedure}.
\newblock \bibinfo{journal}{\emph{arXiv preprint arXiv:2203.05794}} (\bibinfo{year}{2022}).
\newblock


\bibitem[Guo and Berkhahn(2016)]%
        {guo2016entity}
\bibfield{author}{\bibinfo{person}{Cheng Guo} {and} \bibinfo{person}{Felix Berkhahn}.} \bibinfo{year}{2016}\natexlab{}.
\newblock \bibinfo{title}{Entity Embeddings of Categorical Variables}.
\newblock
\newblock
\showeprint[arxiv]{1604.06737}~[cs.LG]


\bibitem[Hogan et~al\mbox{.}(2021)]%
        {Hogan_2021}
\bibfield{author}{\bibinfo{person}{Aidan Hogan}, \bibinfo{person}{Eva Blomqvist}, \bibinfo{person}{Michael Cochez}, \bibinfo{person}{Claudia D’amato}, \bibinfo{person}{Gerard~De Melo}, \bibinfo{person}{Claudio Gutierrez}, \bibinfo{person}{Sabrina Kirrane}, \bibinfo{person}{José Emilio~Labra Gayo}, \bibinfo{person}{Roberto Navigli}, \bibinfo{person}{Sebastian Neumaier}, \bibinfo{person}{Axel-Cyrille~Ngonga Ngomo}, \bibinfo{person}{Axel Polleres}, \bibinfo{person}{Sabbir~M. Rashid}, \bibinfo{person}{Anisa Rula}, \bibinfo{person}{Lukas Schmelzeisen}, \bibinfo{person}{Juan Sequeda}, \bibinfo{person}{Steffen Staab}, {and} \bibinfo{person}{Antoine Zimmermann}.} \bibinfo{year}{2021}\natexlab{}.
\newblock \showarticletitle{Knowledge Graphs}.
\newblock \bibinfo{journal}{\emph{Comput. Surveys}} \bibinfo{volume}{54}, \bibinfo{number}{4} (\bibinfo{date}{July} \bibinfo{year}{2021}), \bibinfo{pages}{1–37}.
\newblock
\showISSN{1557-7341}
\urldef\tempurl%
\url{https://doi.org/10.1145/3447772}
\showDOI{\tempurl}


\bibitem[Hubert et~al\mbox{.}(2022)]%
        {Hubert_2022}
\bibfield{author}{\bibinfo{person}{Nicolas Hubert}, \bibinfo{person}{Armelle Brun}, {and} \bibinfo{person}{Davy Monticolo}.} \bibinfo{year}{2022}\natexlab{}.
\newblock \showarticletitle{New Ontology and Knowledge Graph for University Curriculum Recommendation}.
\newblock


\bibitem[Ji et~al\mbox{.}(2023)]%
        {Ji_2023}
\bibfield{author}{\bibinfo{person}{Ziwei Ji}, \bibinfo{person}{Nayeon Lee}, \bibinfo{person}{Rita Frieske}, \bibinfo{person}{Tiezheng Yu}, \bibinfo{person}{Dan Su}, \bibinfo{person}{Yan Xu}, \bibinfo{person}{Etsuko Ishii}, \bibinfo{person}{Ye~Jin Bang}, \bibinfo{person}{Andrea Madotto}, {and} \bibinfo{person}{Pascale Fung}.} \bibinfo{year}{2023}\natexlab{}.
\newblock \showarticletitle{Survey of Hallucination in Natural Language Generation}.
\newblock \bibinfo{journal}{\emph{Comput. Surveys}} \bibinfo{volume}{55}, \bibinfo{number}{12} (\bibinfo{date}{March} \bibinfo{year}{2023}), \bibinfo{pages}{1–38}.
\newblock
\showISSN{1557-7341}
\urldef\tempurl%
\url{https://doi.org/10.1145/3571730}
\showDOI{\tempurl}


\bibitem[Kang et~al\mbox{.}(2023)]%
        {kang2023knowledge}
\bibfield{author}{\bibinfo{person}{Minki Kang}, \bibinfo{person}{Jin~Myung Kwak}, \bibinfo{person}{Jinheon Baek}, {and} \bibinfo{person}{Sung~Ju Hwang}.} \bibinfo{year}{2023}\natexlab{}.
\newblock \bibinfo{title}{Knowledge Graph-Augmented Language Models for Knowledge-Grounded Dialogue Generation}.
\newblock
\newblock
\showeprint[arxiv]{2305.18846}~[cs.CL]


\bibitem[Karpukhin et~al\mbox{.}(2020)]%
        {karpukhin2020dense}
\bibfield{author}{\bibinfo{person}{Vladimir Karpukhin}, \bibinfo{person}{Barlas Oğuz}, \bibinfo{person}{Sewon Min}, \bibinfo{person}{Patrick Lewis}, \bibinfo{person}{Ledell Wu}, \bibinfo{person}{Sergey Edunov}, \bibinfo{person}{Danqi Chen}, {and} \bibinfo{person}{Wen tau Yih}.} \bibinfo{year}{2020}\natexlab{}.
\newblock \bibinfo{title}{Dense Passage Retrieval for Open-Domain Question Answering}.
\newblock
\newblock
\showeprint[arxiv]{2004.04906}~[cs.CL]


\bibitem[Kim et~al\mbox{.}(2023)]%
        {kim-etal-2023-cot}
\bibfield{author}{\bibinfo{person}{Seungone Kim}, \bibinfo{person}{Se Joo}, \bibinfo{person}{Doyoung Kim}, \bibinfo{person}{Joel Jang}, \bibinfo{person}{Seonghyeon Ye}, \bibinfo{person}{Jamin Shin}, {and} \bibinfo{person}{Minjoon Seo}.} \bibinfo{year}{2023}\natexlab{}.
\newblock \showarticletitle{The {C}o{T} Collection: Improving Zero-shot and Few-shot Learning of Language Models via Chain-of-Thought Fine-Tuning}. In \bibinfo{booktitle}{\emph{Proceedings of the 2023 Conference on Empirical Methods in Natural Language Processing}}, \bibfield{editor}{\bibinfo{person}{Houda Bouamor}, \bibinfo{person}{Juan Pino}, {and} \bibinfo{person}{Kalika Bali}} (Eds.). \bibinfo{publisher}{Association for Computational Linguistics}, \bibinfo{address}{Singapore}, \bibinfo{pages}{12685--12708}.
\newblock
\urldef\tempurl%
\url{https://doi.org/10.18653/v1/2023.emnlp-main.782}
\showDOI{\tempurl}


\bibitem[Lewis et~al\mbox{.}(2019)]%
        {lewis2019bart}
\bibfield{author}{\bibinfo{person}{Mike Lewis}, \bibinfo{person}{Yinhan Liu}, \bibinfo{person}{Naman Goyal}, \bibinfo{person}{Marjan Ghazvininejad}, \bibinfo{person}{Abdelrahman Mohamed}, \bibinfo{person}{Omer Levy}, \bibinfo{person}{Ves Stoyanov}, {and} \bibinfo{person}{Luke Zettlemoyer}.} \bibinfo{year}{2019}\natexlab{}.
\newblock \bibinfo{title}{BART: Denoising Sequence-to-Sequence Pre-training for Natural Language Generation, Translation, and Comprehension}.
\newblock
\newblock
\showeprint[arxiv]{1910.13461}~[cs.CL]


\bibitem[Lewis et~al\mbox{.}(2021)]%
        {lewis2021retrievalaugmented}
\bibfield{author}{\bibinfo{person}{Patrick Lewis}, \bibinfo{person}{Ethan Perez}, \bibinfo{person}{Aleksandra Piktus}, \bibinfo{person}{Fabio Petroni}, \bibinfo{person}{Vladimir Karpukhin}, \bibinfo{person}{Naman Goyal}, \bibinfo{person}{Heinrich Küttler}, \bibinfo{person}{Mike Lewis}, \bibinfo{person}{Wen tau Yih}, \bibinfo{person}{Tim Rocktäschel}, \bibinfo{person}{Sebastian Riedel}, {and} \bibinfo{person}{Douwe Kiela}.} \bibinfo{year}{2021}\natexlab{}.
\newblock \bibinfo{title}{Retrieval-Augmented Generation for Knowledge-Intensive NLP Tasks}.
\newblock
\newblock
\showeprint[arxiv]{2005.11401}~[cs.CL]


\bibitem[Liu et~al\mbox{.}(2021)]%
        {liu2021open}
\bibfield{author}{\bibinfo{person}{Pengfei Liu}, \bibinfo{person}{Youzhang Ning}, \bibinfo{person}{King~Keung Wu}, \bibinfo{person}{Kun Li}, {and} \bibinfo{person}{Helen Meng}.} \bibinfo{year}{2021}\natexlab{}.
\newblock \bibinfo{title}{Open Intent Discovery through Unsupervised Semantic Clustering and Dependency Parsing}.
\newblock
\newblock
\showeprint[arxiv]{2104.12114}~[cs.CL]


\bibitem[McInnes et~al\mbox{.}(2017)]%
        {mcinnes2017hdbscan}
\bibfield{author}{\bibinfo{person}{Leland McInnes}, \bibinfo{person}{John Healy}, \bibinfo{person}{Steve Astels}, {et~al\mbox{.}}} \bibinfo{year}{2017}\natexlab{}.
\newblock \showarticletitle{hdbscan: Hierarchical density based clustering.}
\newblock \bibinfo{journal}{\emph{J. Open Source Softw.}} \bibinfo{volume}{2}, \bibinfo{number}{11} (\bibinfo{year}{2017}), \bibinfo{pages}{205}.
\newblock


\bibitem[Mullick et~al\mbox{.}(2023)]%
        {mullick2023intent}
\bibfield{author}{\bibinfo{person}{Ankan Mullick}, \bibinfo{person}{Ishani Mondal}, \bibinfo{person}{Sourjyadip Ray}, \bibinfo{person}{R Raghav}, \bibinfo{person}{G~Sai Chaitanya}, {and} \bibinfo{person}{Pawan Goyal}.} \bibinfo{year}{2023}\natexlab{}.
\newblock \bibinfo{title}{Intent Identification and Entity Extraction for Healthcare Queries in Indic Languages}.
\newblock
\newblock
\showeprint[arxiv]{2302.09685}~[cs.IR]


\bibitem[Nguyen and Tuan~Nguyen(2020)]%
        {nguyen-tuan-nguyen-2020-phobert}
\bibfield{author}{\bibinfo{person}{Dat~Quoc Nguyen} {and} \bibinfo{person}{Anh Tuan~Nguyen}.} \bibinfo{year}{2020}\natexlab{}.
\newblock \showarticletitle{{P}ho{BERT}: Pre-trained language models for {V}ietnamese}. In \bibinfo{booktitle}{\emph{Findings of the Association for Computational Linguistics: EMNLP 2020}}, \bibfield{editor}{\bibinfo{person}{Trevor Cohn}, \bibinfo{person}{Yulan He}, {and} \bibinfo{person}{Yang Liu}} (Eds.). \bibinfo{publisher}{Association for Computational Linguistics}, \bibinfo{address}{Online}, \bibinfo{pages}{1037--1042}.
\newblock
\urldef\tempurl%
\url{https://doi.org/10.18653/v1/2020.findings-emnlp.92}
\showDOI{\tempurl}


\bibitem[Nguyen and Nguyen(2021)]%
        {phonlp}
\bibfield{author}{\bibinfo{person}{Linh~The Nguyen} {and} \bibinfo{person}{Dat~Quoc Nguyen}.} \bibinfo{year}{2021}\natexlab{}.
\newblock \showarticletitle{{PhoNLP: A joint multi-task learning model for Vietnamese part-of-speech tagging, named entity recognition and dependency parsing}}. In \bibinfo{booktitle}{\emph{Proceedings of the 2021 Conference of the North American Chapter of the Association for Computational Linguistics: Demonstrations}}. \bibinfo{pages}{1--7}.
\newblock


\bibitem[Petroni et~al\mbox{.}(2019)]%
        {petroni2019language}
\bibfield{author}{\bibinfo{person}{Fabio Petroni}, \bibinfo{person}{Tim Rocktäschel}, \bibinfo{person}{Patrick Lewis}, \bibinfo{person}{Anton Bakhtin}, \bibinfo{person}{Yuxiang Wu}, \bibinfo{person}{Alexander~H. Miller}, {and} \bibinfo{person}{Sebastian Riedel}.} \bibinfo{year}{2019}\natexlab{}.
\newblock \bibinfo{title}{Language Models as Knowledge Bases?}
\newblock
\newblock
\showeprint[arxiv]{1909.01066}~[cs.CL]


\bibitem[Radford et~al\mbox{.}(2018)]%
        {radford2018improving}
\bibfield{author}{\bibinfo{person}{Alec Radford}, \bibinfo{person}{Karthik Narasimhan}, \bibinfo{person}{Tim Salimans}, \bibinfo{person}{Ilya Sutskever}, {et~al\mbox{.}}} \bibinfo{year}{2018}\natexlab{}.
\newblock \showarticletitle{Improving language understanding by generative pre-training}.
\newblock  (\bibinfo{year}{2018}).
\newblock


\bibitem[Rizun(2019)]%
        {Rizun_2019}
\bibfield{author}{\bibinfo{person}{Mariia Rizun}.} \bibinfo{year}{2019}\natexlab{}.
\newblock \showarticletitle{Knowledge Graph Application in Education: a Literature Review}.
\newblock \bibinfo{journal}{\emph{Acta Universitatis Lodziensis. Folia Oeconomica}}  \bibinfo{volume}{3} (\bibinfo{date}{08} \bibinfo{year}{2019}), \bibinfo{pages}{7--19}.
\newblock
\urldef\tempurl%
\url{https://doi.org/10.18778/0208-6018.342.01}
\showDOI{\tempurl}


\bibitem[To and Do(2020)]%
        {Huong2020}
\bibfield{author}{\bibinfo{person}{Huong~Duong To} {and} \bibinfo{person}{Phuc Do}.} \bibinfo{year}{2020}\natexlab{}.
\newblock \showarticletitle{Extracting triples from Vietnamese text to create knowledge graph}. In \bibinfo{booktitle}{\emph{12th International Conference on Knowledge and Systems Engineering, {KSE} 2020, Can Tho City, Vietnam, November 12-14, 2020}}. \bibinfo{publisher}{{IEEE}}, \bibinfo{pages}{219--223}.
\newblock
\urldef\tempurl%
\url{https://doi.org/10.1109/KSE50997.2020.9287471}
\showDOI{\tempurl}


\bibitem[Touvron et~al\mbox{.}(2023)]%
        {touvron2023llama2}
\bibfield{author}{\bibinfo{person}{Hugo Touvron}, \bibinfo{person}{Louis Martin}, \bibinfo{person}{Kevin Stone}, \bibinfo{person}{Peter Albert}, \bibinfo{person}{Amjad Almahairi}, \bibinfo{person}{Yasmine Babaei}, \bibinfo{person}{Nikolay Bashlykov}, \bibinfo{person}{Soumya Batra}, \bibinfo{person}{Prajjwal Bhargava}, \bibinfo{person}{Shruti Bhosale}, \bibinfo{person}{Dan Bikel}, \bibinfo{person}{Lukas Blecher}, \bibinfo{person}{Cristian~Canton Ferrer}, \bibinfo{person}{Moya Chen}, \bibinfo{person}{Guillem Cucurull}, \bibinfo{person}{David Esiobu}, \bibinfo{person}{Jude Fernandes}, \bibinfo{person}{Jeremy Fu}, \bibinfo{person}{Wenyin Fu}, \bibinfo{person}{Brian Fuller}, \bibinfo{person}{Cynthia Gao}, \bibinfo{person}{Vedanuj Goswami}, \bibinfo{person}{Naman Goyal}, \bibinfo{person}{Anthony Hartshorn}, \bibinfo{person}{Saghar Hosseini}, \bibinfo{person}{Rui Hou}, \bibinfo{person}{Hakan Inan}, \bibinfo{person}{Marcin Kardas}, \bibinfo{person}{Viktor Kerkez}, \bibinfo{person}{Madian Khabsa},
  \bibinfo{person}{Isabel Kloumann}, \bibinfo{person}{Artem Korenev}, \bibinfo{person}{Punit~Singh Koura}, \bibinfo{person}{Marie-Anne Lachaux}, \bibinfo{person}{Thibaut Lavril}, \bibinfo{person}{Jenya Lee}, \bibinfo{person}{Diana Liskovich}, \bibinfo{person}{Yinghai Lu}, \bibinfo{person}{Yuning Mao}, \bibinfo{person}{Xavier Martinet}, \bibinfo{person}{Todor Mihaylov}, \bibinfo{person}{Pushkar Mishra}, \bibinfo{person}{Igor Molybog}, \bibinfo{person}{Yixin Nie}, \bibinfo{person}{Andrew Poulton}, \bibinfo{person}{Jeremy Reizenstein}, \bibinfo{person}{Rashi Rungta}, \bibinfo{person}{Kalyan Saladi}, \bibinfo{person}{Alan Schelten}, \bibinfo{person}{Ruan Silva}, \bibinfo{person}{Eric~Michael Smith}, \bibinfo{person}{Ranjan Subramanian}, \bibinfo{person}{Xiaoqing~Ellen Tan}, \bibinfo{person}{Binh Tang}, \bibinfo{person}{Ross Taylor}, \bibinfo{person}{Adina Williams}, \bibinfo{person}{Jian~Xiang Kuan}, \bibinfo{person}{Puxin Xu}, \bibinfo{person}{Zheng Yan}, \bibinfo{person}{Iliyan Zarov}, \bibinfo{person}{Yuchen
  Zhang}, \bibinfo{person}{Angela Fan}, \bibinfo{person}{Melanie Kambadur}, \bibinfo{person}{Sharan Narang}, \bibinfo{person}{Aurelien Rodriguez}, \bibinfo{person}{Robert Stojnic}, \bibinfo{person}{Sergey Edunov}, {and} \bibinfo{person}{Thomas Scialom}.} \bibinfo{year}{2023}\natexlab{}.
\newblock \bibinfo{title}{Llama 2: Open Foundation and Fine-Tuned Chat Models}.
\newblock
\newblock
\showeprint[arxiv]{2307.09288}~[cs.CL]


\bibitem[Vedula et~al\mbox{.}(2019)]%
        {vedula2019open}
\bibfield{author}{\bibinfo{person}{Nikhita Vedula}, \bibinfo{person}{Nedim Lipka}, \bibinfo{person}{Pranav Maneriker}, {and} \bibinfo{person}{Srinivasan Parthasarathy}.} \bibinfo{year}{2019}\natexlab{}.
\newblock \bibinfo{title}{Towards Open Intent Discovery for Conversational Text}.
\newblock
\newblock
\showeprint[arxiv]{1904.08524}~[cs.IR]


\bibitem[Vedula et~al\mbox{.}(2020)]%
        {Vedula2020}
\bibfield{author}{\bibinfo{person}{Nikhita Vedula}, \bibinfo{person}{Nedim Lipka}, \bibinfo{person}{Pranav Maneriker}, {and} \bibinfo{person}{Srinivasan Parthasarathy}.} \bibinfo{year}{2020}\natexlab{}.
\newblock \showarticletitle{Open Intent Extraction from Natural Language Interactions}. In \bibinfo{booktitle}{\emph{Proceedings of The Web Conference 2020}} (Taipei, Taiwan) \emph{(\bibinfo{series}{WWW '20})}. \bibinfo{publisher}{Association for Computing Machinery}, \bibinfo{address}{New York, NY, USA}, \bibinfo{pages}{2009–2020}.
\newblock
\showISBNx{9781450370233}
\urldef\tempurl%
\url{https://doi.org/10.1145/3366423.3380268}
\showDOI{\tempurl}


\bibitem[Wang et~al\mbox{.}(2020)]%
        {wang2020minilm}
\bibfield{author}{\bibinfo{person}{Wenhui Wang}, \bibinfo{person}{Furu Wei}, \bibinfo{person}{Li Dong}, \bibinfo{person}{Hangbo Bao}, \bibinfo{person}{Nan Yang}, {and} \bibinfo{person}{Ming Zhou}.} \bibinfo{year}{2020}\natexlab{}.
\newblock \bibinfo{title}{MiniLM: Deep Self-Attention Distillation for Task-Agnostic Compression of Pre-Trained Transformers}.
\newblock
\newblock
\showeprint[arxiv]{2002.10957}~[cs.CL]


\bibitem[Wei et~al\mbox{.}(2022)]%
        {wei2022emergent}
\bibfield{author}{\bibinfo{person}{Jason Wei}, \bibinfo{person}{Yi Tay}, \bibinfo{person}{Rishi Bommasani}, \bibinfo{person}{Colin Raffel}, \bibinfo{person}{Barret Zoph}, \bibinfo{person}{Sebastian Borgeaud}, \bibinfo{person}{Dani Yogatama}, \bibinfo{person}{Maarten Bosma}, \bibinfo{person}{Denny Zhou}, \bibinfo{person}{Donald Metzler}, \bibinfo{person}{Ed~H. Chi}, \bibinfo{person}{Tatsunori Hashimoto}, \bibinfo{person}{Oriol Vinyals}, \bibinfo{person}{Percy Liang}, \bibinfo{person}{Jeff Dean}, {and} \bibinfo{person}{William Fedus}.} \bibinfo{year}{2022}\natexlab{}.
\newblock \bibinfo{title}{Emergent Abilities of Large Language Models}.
\newblock
\newblock
\showeprint[arxiv]{2206.07682}~[cs.CL]


\end{thebibliography}
